\pdfoutput=1

\documentclass[10pt,journal,compsoc]{IEEEtran}

\usepackage[nocompress]{cite}
\usepackage[pdftex]{graphicx}
\usepackage{amsmath}
\usepackage{array}
\usepackage{multirow}
\usepackage{ragged2e}
\usepackage{xifthen}

\usepackage{newpxmath}
\usepackage{microtype}
\usepackage[table]{xcolor}
\usepackage{booktabs}
\usepackage{colortbl}
\usepackage{tikz}
\usepackage{csquotes}
\usepackage[inline]{enumitem}
\usepackage{xparse}
\usepackage{pifont}
\usepackage{wrapfig}
\usepackage[hidelinks]{hyperref}

\input{confusionflow-macros.tex}

\begin{document}

\title{ConfusionFlow: A model-agnostic visualization for
temporal analysis of classifier confusion}

\author{%
    Andreas~Hinterreiter\(^{*}\),
    Peter~Ruch\(^{*}\),
    Holger~Stitz,
    Martin~Ennemoser,\\
    J\"urgen~Bernard,
    Hendrik~Strobelt,
    and~Marc~Streit
    \IEEEcompsocitemizethanks{%
        \IEEEcompsocthanksitem Andreas Hinterreiter, Peter Ruch, Holger Stitz,
        and Marc Streit are with Johannes Kepler University Linz.\newline
        E-mail: \{andreas.hinterreiter, peter.ruch, holger.stitz, marc.streit\}@jku.at
        \IEEEcompsocthanksitem Andreas Hinterreiter is also with Imperial College London.\newline
        Email: \{a.hinterreiter@imperial.ac.uk\}
        \IEEEcompsocthanksitem Martin Ennemoser is with Salesbeat GmbH.\newline
        Email: m.ennemoser@salesbeat.io
        \IEEEcompsocthanksitem Hendrik Strobelt is with IBM Research.\newline
        E-mail: hendrik.strobelt@ibm.com.
        \IEEEcompsocthanksitem J\"urgen Bernard is with The University of
        British Columbia.\newline
        E-mail: jubernar@cs.ubc.ca.
        \IEEEcompsocthanksitem[\(*\)] These authors contributed equally to this work.%
    }%
    \thanks{Manuscript received XX Month 2019.}%
}

\markboth{IEEE Transactions on Visualization and Computer Graphics}%
{IEEE Transactions on Visualization and Computer Graphics}

\IEEEtitleabstractindextext{%
\begin{abstract}
    Classifiers are among the most widely used supervised machine learning algorithms.
    Many classification models exist, and choosing the right one for a given task is difficult.
    During model selection and debugging, data scientists need to assess classifier\CFinsert{s'} performance\CFinsert{s}, evaluate the\CFinsert{ir} \CFreplace{train}{learn}ing behavior over time, and compare different models.
    Typically, this analysis is based on single-number performance measures such as accuracy.
    A more detailed evaluation of classifiers is possible by inspecting class errors.
    The confusion matrix is an established way for visualizing these class errors, but it was not designed with temporal or comparative analysis in mind.
    More generally, established performance analysis systems do not allow a combined temporal and comparative analysis of class-level information.
    To address this issue, we propose ConfusionFlow, an interactive, comparative visualization tool that combines the benefits of class confusion matrices with the visualization of performance characteristics over time.
    ConfusionFlow is model-agnostic and can be used to compare performances for different model types, model architectures, and/or training and test datasets.
    We demonstrate the usefulness of ConfusionFlow in a case study on instance selection strategies in active learning.
    We further assess \CFinsert{the} scalability \CFreplace{issues and possible mitigation mechanisms when }{of} ConfusionFlow \CFreplace{is applied to problems with a high number of classes}{and present a use case in the context of neural network pruning}.
\end{abstract}
%
}

\maketitle

%
\IEEEpeerreviewmaketitle

\IEEEraisesectionheading{\section{Introduction}\label{sec:introduction}}

\IEEEPARstart{C}{lassification} is one of the most frequent machine learning (ML) tasks.
Many important problems from diverse domains, such as image processing~\cite{krizhevsky_imagenet_2017, he_deep_2016}, natural language processing~\cite{nogueira_dos_santos_deep_2014, socher_recursive_2013, glorot_domain_2011}, or drug target prediction~\cite{mayr_large-scale_2018}, can be framed as classification tasks.
Sophisticated models, such as neural networks, have been proven to be effective, but building and applying these  models is difficult.
This is especially the case for multiclass classifiers, which can predict one out of several classes---as opposed to binary classifiers, which can only predict one out of two.

During \CFinsert{the} development of classifiers, data scientists are confronted with a series of challenges.
They need to observe how the model performance develops over time, where the notion of \emph{time} can be twofold: on the one hand, the general workflow in ML development is incremental and iterative, typically consisting of\CFdelete{ testing} many \CFinsert{sequential experiments with} different models\CFdelete{ sequentially}; on the other hand, the actual (algorithmic) training of a classifier is itself an optimization problem, involving different model states over time.
In the first case, comparative analysis helps the data scientists gauge whether they are on the right track.
In the latter case, temporal analysis helps to find the right time to stop the training, \CFreplace{such}{so} that the model generalizes to unseen samples not represented in the training data.

Model behavior can depend strongly on the choice\CFinsert{s} of hyperparameters, optimizer, or loss function.
It is usually not obvious how these choices affect the overall model performance.
It is even less obvious how these choices might affect the behavior on a more detailed level, such as commonly \enquote{confused} pairs of classes.
However, knowledge about the class-wise performance of models can help data scientists make more informed decisions.

To cope with these challenges, data scientists employ three different types of approaches.
One, they assess single value performance measures such as accuracy, typically by looking at temporal line charts.
This approach is suitable for comparing learning behavior, but it inherently lacks information at the more detailed class level.
Two, data scientists use tools for compari\CFinsert{ng the}\CFdelete{son of classifier} performance\CFinsert{s of classifiers}.
However, these tools typically suffer from the same lack of class-level information, or they are not particularly suited for temporal analysis.
Three, data scientists assess class-level performance from the class-confusion matrix~\cite{sokolova_systematic_2009}.
Unfortunately, the rigid layout of the classic confusion matrix does not lend itself well to model comparison or temporal analysis.

So far, few tools have focused on classification analysis from a \CFinsert{combined} temporal, model-comparative, and class-level \CFreplace{point of view}{perspective}.
However, \CFreplace{data scientists could benefit from such a temporal-comparative assessment of classifier performance by} gaining insights from all three points of view in a single tool\CFdelete{.
Especially for training neural networks, this additional feedback} can \CFinsert{(1)}~serve as a starting point \CFreplace{to help data scientists}{for} interpret\CFinsert{ing} \CFreplace{the} model performance\CFinsert{s}, \CFinsert{(2)}~facilitate \CFinsert{the} navigati\CFreplace{ng}{ion through} the space of \CFreplace{possible adaptions of the model}{model adaptations}, and \CFinsert{(3)}~\CFreplace{improve}{lead to a better} understanding of the interaction\CFinsert{s} \CFreplace{of the}{between a} model and \CFinsert{the} underlying data.

\begin{figure*}
    \centering
    \includegraphics[width=\textwidth]{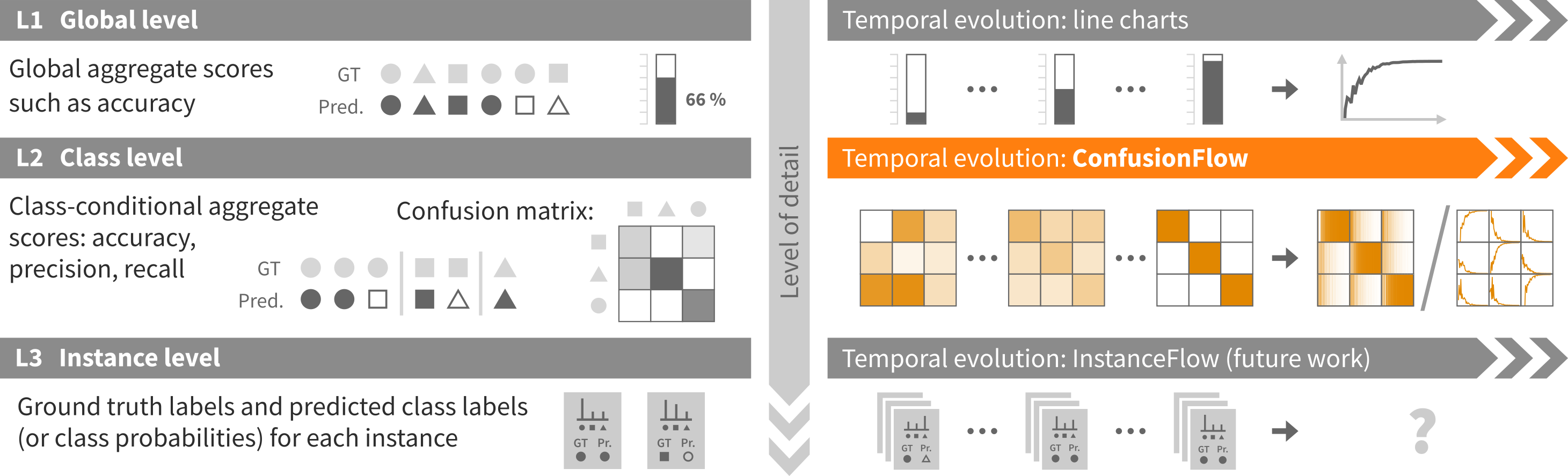}
    \caption{A classifier's performance can be evaluated on three \CFdelete{different }levels of detail: global aggregate scores~(\ref{lvl:global}); class-conditional scores and class confusion~(\ref{lvl:class}); and detailed, instance-wise information~(\ref{lvl:instance}).
    ConfusionFlow operates on the class level \ref{lvl:class}, \CFreplace{concentrating on the}{enabling a} temporal analysis of \CFinsert{the} \CFreplace{train}{learn}ing behavior.}
    \label{fig:perf-levels}
\end{figure*}

The \textbf{primary contribution} of our paper is \emph{ConfusionFlow}, a precision- and recall-based visualization that \CFdelete{lets data scientists assess classifier performance.
ConfusionFlow }enables temporal, comparative, and class-level performance analysis \emph{at the same time}\CFreplace{,}{.} \CFinsert{To this end, we} introduc\CFreplace{ing}{e} a temporal adaptation of the \CFreplace{classic}{traditional} confusion matrix.

As \textbf{secondary contributions} we present \CFinsert{(1)}~a thorough problem space characterization of classifier performance analysis, including a three-level granularity scheme; \CFinsert{(2)}~a case study showing how ConfusionFlow can be applied to analyze labeling strategies in active learning; \CFreplace{and}{(3)}~an evaluation of \CFinsert{ConfusionFlow's} scalability\CFreplace{ issues and possible mitigation mechanisms for confusion matrices in the case of many classes}{; and (4)~a usage scenario in the context of neural network pruning}.

\section{Problem Space Characterization}
\label{sec:tasks}

The development of new classification models or the adaptation of an existing model to a new application domain are highly experimental tasks.
A user is confronted with many different design decisions, such as choosing an architecture, a suitable optimization method, and a set of hyperparameters.
All of these choices \CFdelete{considerably }affect the learning behavior \CFinsert{considerably,} and \CFdelete{ultimately}{influence} the quality of the final classifier.
Consequently, to obtain satisfying results, multiple classifiers based on different models or configurations need to be trained and evaluated in an iterative workflow.
Here, we chose the term \emph{configuration} to refer to the set of model, optimization techniques, hyperparameters, and input data.
This design process requires the user to compare the learning behavior\CFinsert{s} and performances \CFreplace{for}{of} different models or configurations across multiple training iterations\CFreplace{, for instance by inspecting}{. To this end, model developers typically inspect} performance measures such as precision and recall\CFinsert{\footnote{We provide detailed definitions of the most common performance metrics in the Supplementary Information (see Section S\,1.}}.
\CFdelete{We provide detailed definitions of the most common performance metrics in the Supplementary Information.}
\CFreplace{This performance}{Depending on the measures used, the} analysis can be carried out on three \CFdelete{different }levels of detail.

\subsection{Analysis Granularity}
\label{sec:levels}

Based on our reflection of related works (see Section~\ref{sec:related-work}), most performance analysis tasks for classifiers can be carried out on three \CFdelete{different }levels of detail (see Figure~\ref{fig:perf-levels}, left):

\begin{levels}
    \item[Global level]%
    \label{lvl:global}%
        At the global level, the classifier's performance is judged by aggregate scores that sum up the results for the entire dataset in a single number.
        The overall accuracy is a typical example for a global performance measure.
        For showing trends across multiple training iterations, global aggregate scores can be represented in simple line charts.
    \item[Class level]%
    \label{lvl:class}%
        At the class level, performance measures are derived from subsets of the results based on specific class labels.
        Typical performance measures at the class level are class-wise accuracy, precision, or recall.
        Like for the global level, the temporal evolution of these measures throughout training can be visualized as line charts.
        More detailed class-level information is contained in the confusion matrix.
        This \CFreplace{paper}{work} addresses the problem of visualizing the confusion matrix across multiple training iterations.
    \item[Instance level]%
    \label{lvl:instance}%
        At the instance level, quality assessment is based on individual ground truth labels and predicted labels (or predicted class probabilities). This allows picking out problematic input data.
        Strategies for how to further analyze these instances vary strongly between different models and data types.
        Depending on the specific problem, interesting information may be contained in input images or feature vectors, network outputs, neuron activations, and/or more advanced concepts such as saliency maps~\cite{simonyan_deep_2013}.
\end{levels}

These three levels are different degrees of aggregation of individual instance predictions.
Figure~\ref{fig:perf-levels} (right) shows schematically how data at these levels can be visualized across iterations to enable \CFinsert{an} analysis of the training progress.

ConfusionFlow aims at \CFinsert{a} temporal analysis of per-class aggregate scores, introducing a visualization of the confusion matrix across training iterations.
ConfusionFlow thus operates on the second of the three levels~(\ref{lvl:class}).

\subsection{Analysis Tasks}
\label{sec:comparison-and-temporal}

ConfusionFlow is designed for experts in data science and ML, ranging from model developers to model users.
Building upon the reflection of these user groups from related work (see Section~\ref{sec:related-work}) and the characterization of the working practices of our collaborators, we break-down user goals and intents into a task characterization as follows.
The principal structure of tasks is along two \CFdelete{cross-cutting }axes\CFreplace{.
These two axes}{, which} correspond to the two high-level goals of \emph{comparative} analysis (\ref{hl-task:comp}) and \emph{temporal} analysis (\ref{hl-task:temp}).

The comparative axis differentiates \emph{within}-classification comparisons from \emph{between}-classification comparisons.
This frequently used \CFreplace{theme of within- and between-analyses}{within\slash between dichotomy} accounts for analyses that are conducted with a single classification result (within), as well as those relevant for multiple results (between).
The notion of within-comparison also alludes to the different levels of detail discussed in Section~\ref{sec:levels}.
All between-classification analyses share the principle of comparing multiple classification results.
In general, users have a series of classification configurations at hand, aiming at identifying commonalities and differences.
According to our collaborators, multiple classification configurations can result from 
\begin{enumerate*}[
    label={(\alph*)},
    itemjoin={{, }},
    itemjoin*={{, or }},
    after={.}]
\item different (hyper-)parameter choices of the same model
\item different classification models
\item different datasets or dataset folds used for classifier training and validation
\end{enumerate*}
While the differentiation between these three different types of classification configurations is interesting from a methodological perspective, the requirements to tools for quality assessment are very similar.

The second axis structures \CFinsert{the} temporal analysis tasks (\ref{hl-task:temp}).
Along this axis\CFdelete{,} we differentiate between tasks as they \CFinsert{are} typically applied in time series analysis~\cite{aigner2011visualization}: looking up values, assessing trends, and finding anomalies.

The complete crosscut of these two axes leads to six primary analysis tasks, \ref{task:measure-perf} to~\ref{task:relate-problem}, which are listed in Table~\ref{tab:tasks}.
Fore each task, an exemplary scenario is given, which explains how the abstract low-level task relates to the ML analysis procedure.

In section~\ref{sec:levels}\CFdelete{ on the levels of detail}, we already hinted at the fact that existing tools mostly support temporal analysis~(\ref{hl-task:temp}) only on a global level~(\ref{lvl:global}).
As our literature survey below will show, comparison between classifiers~(\ref{hl-task:comp-between}) is \CFdelete{also }rarely supported on a class level.
The main novelty of ConfusionFlow is enabling precision- and recall-based class-level analysis in a comparative \emph{and} temporal way.

\begin{table*}
    \caption{Analysis tasks relevant to the design of ConfusionFlow.
    The space of low-level tasks \ref{task:measure-perf}--\ref{task:relate-problem} is generated by two axes of high level goals, comparative~(\ref{hl-task:comp}) and temporal analysis~(\ref{hl-task:temp}), respectively.}
    \label{tab:tasks}
    \centering
    \footnotesize\sffamily
    \setlength{\extrarowheight}{.5ex}
    \setlength{\aboverulesep}{0pt}
    \setlength{\belowrulesep}{0pt}
    \newcolumntype{L}[1]{>{\RaggedRight\arraybackslash}m{#1}}
    \begin{tabular}{%
        @{}r@{~}l@{\quad~}%
        r@{~}L{.2\textwidth}@{\qquad}%
        r@{~}L{.14\textwidth}%
             L{.2\textwidth}%
             L{.14\textwidth}%
        }
        \toprule
        & & \hltask{G1}\label{hl-task:comp} & \multicolumn{5}{@{~}l}{Comparative analysis} \\
        \cmidrule{3-8}
        & & \hltask[G1a]{a}\label{hl-task:comp-within} & Within-classific. comparison &
            \hltask[G1b]{b}\label{hl-task:comp-between} & \multicolumn{3}{@{}l}{Between-classification comparison} \\
        \hltask{G2}\label{hl-task:temp} & Temporal analysis & & & &
            \color{gray!70}Different models & 
                \color{gray!70}Different hyperparameters &
                \color{gray!70}Different data\\
        \cmidrule(r{7pt}){1-2} \cmidrule(r{7pt}){3-4} \cmidrule{5-8}
        \hltask[G2a]{a}\label{hl-task:temp-lookup} & Lookup values &
            \task{T1}\label{task:measure-perf} & Measure performance & 
            \task{T4}\label{task:comp-perf} & \multicolumn{3}{@{}l}{Compare performances} \\
         & & &
            \color{gray!70}Read off quality measure at certain epoch & &
            \color{gray!70}Assess final and intermediate model suitability &
            \color{gray!70}Relate final performance to hyperparameter space &
            \color{gray!70}Estimate final generalization capabilities \\
        \cmidrule(r{7pt}){3-4} \cmidrule{5-8}
        \hltask[G2b]{b}\label{hl-task:temp-trends} & Assess trends &
            \task{T2}\label{task:measure-prog} & Measure learning progress &
            \task{T5}\label{task:comp-prog} & \multicolumn{3}{@{}l}{Compare learning progress} \\
        & & &
            \color{gray!70}Assess characteristic saturation curve for learning process & &
            \color{gray!70}Compare learning speeds for different models &
            \color{gray!70}Relate learning speeds to hyperparameter choices &
            \color{gray!70}Identify over- or underfitting \\
        \cmidrule(r{7pt}){3-4} \cmidrule{5-8}
        \hltask[G2c]{c}\label{hl-task:temp-anom} & Find anomalies &
            \task{T3}\label{task:detect-problem} & Detect temporal problems &
            \task{T6}\label{task:relate-problem} & \multicolumn{3}{@{}l}{Relate temporal problems} \\
        & & &
            \color{gray!70}Identify performance spikes and drops & & 
            \color{gray!70}Relate anomalies to model &
            \color{gray!70}Relate learning failure to parameter choice &
            \color{gray!70}Identify problematic instance sampling \\
        \bottomrule
    \end{tabular}
\end{table*}

\begin{figure}
    \centering
    \includegraphics[width=.9\linewidth]{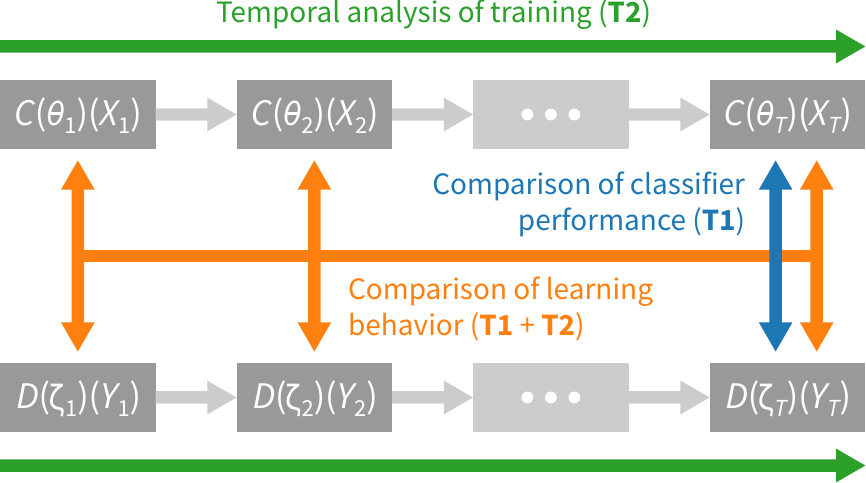}
    \caption{
    Principle working practice as observed with our collaborators.
    The two main goals of comparative and/or temporal analysis of classifiers led us to define the two axes that help to structure the analysis tasks.
    This schematic shows the most general case of different models and different, changing datasets.
    }
    \label{fig:tasks}
\end{figure}

Figure~\ref{fig:tasks} illustrates all possible scenarios of performing the high level tasks \ref{hl-task:comp} and \ref{hl-task:temp} at the same time, with special focus on between-classification comparison~(\ref{hl-task:comp-between}).
This schematic shows the most general case, i.e.,~when both the models (\(C\Leftrightarrow D\)) and the datasets (\(X\Leftrightarrow Y\)) are different, and the datasets additionally change over time (\(X_i\neq X_j\)).
Many specialized comparison and/or temporal analysis tasks can be derived from the general case depicted in Figure~\ref{fig:tasks}, when either the dataset or the model are kept constant:
\begin{itemize}
    \item In the simple case of observing how a single model is trained on a constant data set (\(X_t=X\) for all~\(t\)), the user is interested only in the sequence \(C(\theta_1)(X) \cdots C(\theta_T)(X)\).
    This corresponds to the performance measuring~(\ref{task:measure-perf}) and progress measuring tasks~(\ref{task:measure-prog}).
    \item For comparing the final performances of two classification models, \(C\) and \(D\), acting on the same test set \(X\), the user analyzes the pair \(C(\theta_T)(X)\) vs.~\(D(\zeta_T)(X)\).
    This is a typical realization of task~\ref{task:comp-perf}.
    \item Often, the performances of a classifier on two different dataset folds, such as training and test folds, need to be compared temporally.
    This scenario implies \(C=D\), and \(X_t=X\) and \(Y_t=Y\) for all \(t\), but \(X\neq Y\).
    The user now needs to compare the sequence \(C(\theta_1)(X) \cdots  C(\theta_T)(X)\) with the sequence \(C(\theta_1)(Y) \cdots C(\theta_T)(Y)\).
    This analysis includes \CFdelete{performing} a comparative \CFinsert{assessment of} trend\CFinsert{s} and anomal\CFreplace{y assessment}{ies}  (tasks~\ref{task:comp-prog} and~\ref{task:relate-problem}).
    \item A more complex example is the comparison of two classifiers during active learning (see Section~\ref{sec:case-study-active-learning}).
    In this case, both models are trained on the same dataset, but the dataset changes over time.
    The user \CFdelete{then }compares the sequence \(C(\theta_1)(X_1) \cdots  C(\theta_T)(X_T)\) with the sequence \(D(\zeta_1)(X_1) \cdots  D(\zeta_T)(X_T)\).
    All fine-grained tasks \ref{task:measure-perf} to~\ref{task:relate-problem} may be relevant in this complex example.
\end{itemize}

We \CFreplace{would like to stress that we}{deliberately} kept all example scenarios in Table~\ref{tab:tasks} level-agnostic.
The tasks \ref{task:measure-perf} to~\ref{task:relate-problem} are equally relevant on any of the three levels of detail.
ConfusionFlow focuses on enabling users to perform the tasks on the class level \ref{lvl:class}, but also includes some global information deducible from the confusion matrix.

\section{Related Work}
\label{sec:related-work}

The recent resurgence of ML in general and the increasing popularity of deep learning in particular have led to an increased demand for ML development and monitoring tools, but also to an increased desire to better understand existing techniques.
This interplay between algorithm design on the one hand, and the challenge of making ML algorithms explainable or interpretable on the other hand, has spawned high activity in the field of visualization for ML.
The interrogative survey by Hohman et al.~\cite{hohman_visual_2018} gives a comprehensive overview of these recent advances.

In the following, we will discuss approaches that
\begin{enumerate*}[
    label={(\arabic*)},
    itemjoin={{; }},
    itemjoin*={{; and/or }},
    after={.}]
\item target the user goal of comparison across models and/or configurations (see goal~\ref{hl-task:comp-between})
\item enable a temporal analysis of the \CFreplace{train}{learn}ing behavior (see goal \ref{hl-task:temp})
\item operate on the class level (\ref{lvl:class}) as defined in our contextualization in Section~\ref{sec:levels}
\end{enumerate*}
Table~\ref{tab:related-work} summarizes which of these three aspects is covered by each of our selected publications.
\CFreplace{As}{We also briefly review  previous work on time series visualization and comparison, since} ConfusionFlow is a small multiples \CFreplace{visualization of time series and}{approach, that} should support users in performing typical temporal analysis tasks\CFinsert{.}\CFdelete{, we also briefly review previous work in the field of time series visualization.}%

Our literature survey will show that hardly any tool so far has focused on simultaneously addressing---on the class level of detail~(\ref{lvl:class})---the two high-level goals of comparison (especially between models, \ref{hl-task:comp-between}) and temporal analysis~(\ref{hl-task:temp}).

\subsection{Model Comparison}

\CFinsert{Gleicher et al.~structured the design space for visual comparisons into three main categories: juxtaposition, superposition, and explicit representation~\cite{gleicher_visual_2011}.
Most ML model comparison visualizations use superposition for plots of single-number performance metrics, and juxtaposition for comparison of multidimensional (e.g., vectors or matrices) or unstructured data (e.g., images or text).}

One of the most well-known visualization systems for developing, debugging, and evaluating neural networks is TensorFlow's TensorBoard by Abadi et al.~\cite{abadi_tensorflow:_2016, wongsuphasawat_visualizing_2018}.
It combines a visualization of the computation graph with a display of various performance metrics, but it is not designed for comparing multiple ML models in the same view.
TensorFlow also features Google's What-If Tool~\cite{google_whatif}, which enables comparison of a model with a changed version of itself upon hypothetical changes to the dataset.

For certain types of model architecture, tools with specialized comparison features have been developed: RNNVis by Ming et al.~\cite{ming_understanding_2017} for recurrent neural networks\CFreplace{;}{,} GANViz by Wang et al.~\cite{wang_ganviz:_2018} for generative adversarial networks\CFreplace{;}{,} and CNNComparator by Zeng et al.~\cite{zeng_cnncomparator:_2017} for convolutional neural networks.
RNNVis features a main view with a glyph based sentence visualization.
On demand, two models can be compared side by side.
GANViz focuses on the comparison of the outputs of the generative network with those of the discriminative network that together make up the GAN.
CNNComparator consists of a histogram of parameter values for a selected network layer, a matrix visualization of the convolution operation, as well as an instance-level side-by-side comparison of two selected networks' performances.
It allows comparison of two different configurations or of two different model states for the same configuration, but does not feature immediate access to class confusion measures.
ShapeShop by Hohman et al.~\cite{hohman_shapeshop:_2017} is aimed at non-experts, and enables comparison of the performances of convolutional neural networks.
It is designed to \CFreplace{support}{give} the user \CFreplace{in}{a basic} understanding \CFdelete{the basics }of what the network learns, rather than provide in-depth evaluation functionality.

Zhang et al.~presented Manifold~\cite{zhang_manifold:_2019}, a model-agnostic framework for interpreting, comparing, and diagnosing ML models.
Small multiples of scatter plots visualize how two different models generate different class outputs.
Color coding gives a sense of each model's class confusion, but there is no option to track the models' \CFreplace{train}{learn}ing behaviors.

Comparison of two models cannot only be used to select which model performs better on its own.
It can also be part of a workflow to construct new ensemble models or adapt models interactively.
In van den Elzen's and van Wijk's BaobabView~\cite{van_den_elzen_baobabview:_2011}, decision trees can be pruned interactively, and the performances of the resulting tree can be compared to the initial one, e.g., by looking at the confusion matrices.
EnsembleMatrix by Talbot et al.~\cite{talbot_ensemblematrix:_2009} displays confusion matrices for different classifiers, allowing the user to construct a weighted combination from among them.
The resulting ensemble model can be evaluated, again in terms of class confusion.

Each of these techniques enables the user to compare the performance\CFinsert{s} of multiple models or model states in some way (addressing goal~\ref{hl-task:comp-between}), but misses either the temporal aspect~(\ref{hl-task:temp}), or does not yield class confusion information~(\ref{lvl:class}).

\subsection{Temporal Analysis of Training}

We subdivide the review of temporal analysis of model training into an elaboration of relevant temporal analysis tasks, followed by an overview of approaches supporting the temporal assessment of quality characteristics of classifiers.

In data mining, ML, and visualization research, \CFdelete{a }principal goal\CFinsert{s} of temporal data analysis \CFreplace{is}{are} the exploration, identification, \CFreplace{or}{and} localization of temporal patterns~\cite{fayyad1996data,FU2011164,MIKSCH2014286}.
From task characterizations in the visualization community, we adopt the notion of localizing single (i.e., atomic) values as the finest granularity of temporal analysis~\cite{andrienko2006exploratory,aigner2011visualization}.
Building upon these atomic localization tasks, most visualization techniques support the assessment of multiple points in time, enabling users to grasp higher-level structures, i.e., temporal patterns.
Depending on the application context, such patterns often include trends, outliers, or periodic patterns~\cite{bernard2015}.
For the temporal analysis of model training, the identification of trends plays a crucial role.
As a rule, model performances tend to increase throughout the learning process converging towards some saturation value.
\CFreplace{The second type of pattern that is}{Other} frequently investigated \CFinsert{patterns} include\CFdelete{s} outliers \CFreplace{or}{and} anomalies.
\CFreplace{On the one hand, outliers and anomalies are literally difficult to characterize, but on the other hand are highly relevant to be assessed.}{The assessment of anomalies in the context of ML models is highly relevant but challenging.}
In contrast to trends and outliers, periodic patterns and cycles hardly \CFreplace{occur}{play a role} in the temporal analysis of classifier training.

We \CFdelete{will }now give an overview of specific ML performance analysis tools that have addressed these general temporal analysis tasks.

TensorBoard~\cite{abadi_tensorflow:_2016} and GanViz~\cite{wang_ganviz:_2018} augment their main visualization with line charts of accuracy or other performance scores.
Similarly, Chung et al.~\cite{chung_revacnn:_2016} show temporal training statistics in an extra window of their ReVACNN system for real-time analysis of convolutional neural networks.
In CNNComparator~\cite{zeng_cnncomparator:_2017}, limited temporal information is accessible by comparing two model states from different training epochs.

DeepEyes by Pezzotti et al.~\cite{pezzotti_deepeyes:_2018} is a progressive visualization tool that combines curves for loss and accuracy with perplexity histograms and activation maps.
Progressive line charts of loss during the training are also used in educational tools for interactive exploration, such as TensorFlow Playground~\cite{smilkov_direct-manipulation_2017} or GAN Lab~\cite{kahng_gan_2019}.

DeepTracker by Liu et al.~\cite{liu_deeptracker:_2018} displays performance data in a cube-style visualization, where training epochs progress along one of the three axes.
A different approach to enable inspection of the \CFreplace{train}{learn}ing behavior is a selector or slider \CFreplace{that allows accessing individual iterations, and that}{which} is linked to a main visualization or multiple visualizations in a dashboard \CFinsert{and allows accessing individual iterations}.
Chae et al.~\cite{chae_visualization_2017} made use of this technique in their visualization of classification outcomes, as did Wang et al.~in DQNViz~\cite{wang_dqnviz:_2019}, a tool for understanding Deep Q-networks.
In one of the views in Bruckner's ML-o-scope~\cite{bruckner_ml-o-scope:_2014} an epoch slider is tied to a confusion matrix, in which cells can be interactively augmented with example instances.
The Blocks system by Alsallakh et al.~\cite{alsallakh_convolutional_2018} also features a confusion matrix bound to an epoch slider.
Blocks supports investigation of a potential class hierarchy learned by neural networks, which requires the visualization to be scalable to many classes.

Of all the tools for exploration of the \CFreplace{train}{learn}ing behavior~(\ref{hl-task:temp}) mentioned above, none focuses on class confusion~(\ref{lvl:class}) while also providing comparison functionality~(\ref{hl-task:comp-between}).

\begin{table}
    \caption{Publications related to ConfusionFlow, classified by whether they allow between\slash classification comparison~(\ref{hl-task:comp-between}), offer temporal information~(\ref{hl-task:temp}), and/or operate at the class level~(\ref{lvl:class}).}
    \label{tab:related-work}
    \centering
    \footnotesize\sffamily
    \newcommand{\yes}{\textcolor{CFgreen!80!black}{\ding{52}}}
    \newcommand{\meh}{\textcolor{gray!50}{\ding{52}}}
    \newcommand{\no}{}
    \newcommand{\na}{\textcolor{gray!70}{N/A}}
    \newcommand{\noname}{\textcolor{gray!70}{unnamed}}
    \rowcolors{2}{}{lightgray!15}
    \setlength{\extrarowheight}{.5ex}
    \setlength{\aboverulesep}{0pt}
    \setlength{\belowrulesep}{0pt}
    \begin{tabular}{ll*{3}c}
        System & Publication  & 
        \ref{hl-task:comp-between} & 
        \ref{hl-task:temp} & 
        \ref{lvl:class}  \\
        \midrule
        RNNVis      & Ming et al.~\cite{ming_understanding_2017}    & \yes & \no  & \na  \\
        CNNComparator & Zeng et al.~\cite{zeng_cnncomparator:_2017} & \yes & \meh & \no  \\
        GANViz      & Wang et al.~\cite{wang_ganviz:_2018}          & \yes & \meh & \meh \\
        DQNViz      & Wang et al.~\cite{wang_dqnviz:_2019}          & \yes & \yes & \na  \\
        TensorBoard & Abadi et al.~\cite{abadi_tensorflow:_2016}    & \meh & \yes & \no  \\
        ReVACNN     & Chung et al.~\cite{chung_revacnn:_2016}       & \no  & \yes & \no  \\
        DeepEyes    & Pezzotti et al.~\cite{pezzotti_deepeyes:_2018}& \no  & \yes & \no  \\
        DeepTracker & Liu et al.~\cite{liu_deeptracker:_2018}       & \no  & \yes & \meh \\
        \noname     & Chae et al.~\cite{chae_visualization_2017}    & \meh & \yes & \meh \\
        Blocks      & Alsallakh et al.~%
                            \cite{alsallakh_convolutional_2018}     & \no  & \yes & \yes \\
        ML-o-scope  & Bruckner~\cite{bruckner_ml-o-scope:_2014}     & \no  & \yes & \yes \\
        Confusion wheel & Alsallakh et al.~%
                            \cite{alsallakh_visual_2014}            & \no  & \no  & \yes \\
        ManiMatrix  & Kapoor et al.~\cite{kapoor_interactive_2010}  & \no  & \no  & \yes \\
        Squares     & Ren et al.~\cite{ren_squares:_2017}           & \meh & \no  & \yes \\
        BaobabView  & v.\,d. Elzen \& v. Wijk~%
                            \cite{van_den_elzen_baobabview:_2011}   & \yes & \no  & \yes \\
        EnsembleMatrix & Talbot et al.~%
                            \cite{talbot_ensemblematrix:_2009}      & \yes & \no  & \yes \\
        Manifold    & Zhang et al.~\cite{zhang_manifold:_2019}      & \yes & \no  & \yes \\
        \midrule
        \multicolumn{5}{l}{%
            \yes~Covered\quad
            \meh~Partly covered\quad
            \na~Not applicable}
    \end{tabular}
\end{table}

\subsection{Class Confusion}

When evaluating the output of classifiers at level \ref{lvl:class}, class confusion can be interpreted in two ways.
Typically, it describes the aggregate scores used in the individual cells of the confusion matrix.
However, the term \enquote{between-class confusion} is sometimes also used to describe high probability values for more than one class in a classifier's output for an individual instance.
In order to avoid ambiguity, we will call this notion \enquote{per-instance classification uncertainty} in our discussion.

Of the works mentioned so far, BaobabView~\cite{van_den_elzen_baobabview:_2011}, EnsembleMatrix~\cite{talbot_ensemblematrix:_2009}, ML-o-scope~\cite{bruckner_ml-o-scope:_2014}, and Blocks~\cite{alsallakh_convolutional_2018} all allow, at least partially, performance analysis on the class level~(\ref{lvl:class}).
In these tools, this is realized by visualizing class confusion in terms of standard confusion matrices, either for the final classifier or for one training step at a time.

The confusion matrix is also the heart of the ManiMatrix tool by Kapoor et al.~\cite{kapoor_interactive_2010}, where it is used to interactively modify classification boundaries.
This lets the user explore how constraining the confusion for one pair of classes affects the other pairs, aiming at class-level model optimization and interpretability.

Next to the confusion matrix, some alternative ways of evaluating classifier performance on level \ref{lvl:class} have been proposed.
Alsallakh et al.~introduced the confusion wheel~\cite{alsallakh_visual_2014}.
It consists of a circular chord diagram, in which pairs of classes with high confusion counts are connected with thicker chords.
On the outside, ring charts encode FN, FP, TP, and TN distributions for each class.
Squares by Ren et al.~\cite{ren_squares:_2017} is focused on visualizing per-instance classification uncertainty.
Histograms of prediction scores can be unfolded to access individual instances, whose predictions are then encoded using parallel coordinates.
Additionally, sparklines for each class give an impression of aggregate class confusion.
Squares allows \CFinsert{a} hybrid-level~(\ref{lvl:class} and \ref{lvl:instance}) confusion analysis.

None of the existing tools for class-level performance analysis~(\ref{lvl:class}) provide an immediate, temporal representation of the \CFreplace{train}{learn}ing behavior~(\ref{hl-task:temp}), and most are relatively ill-suited for between-classification comparison~(\ref{hl-task:comp-between}).

\section{ConfusionFlow Technique}
\label{sec:technique}

\begin{figure*}
    \centering
    \includegraphics[width=\textwidth]{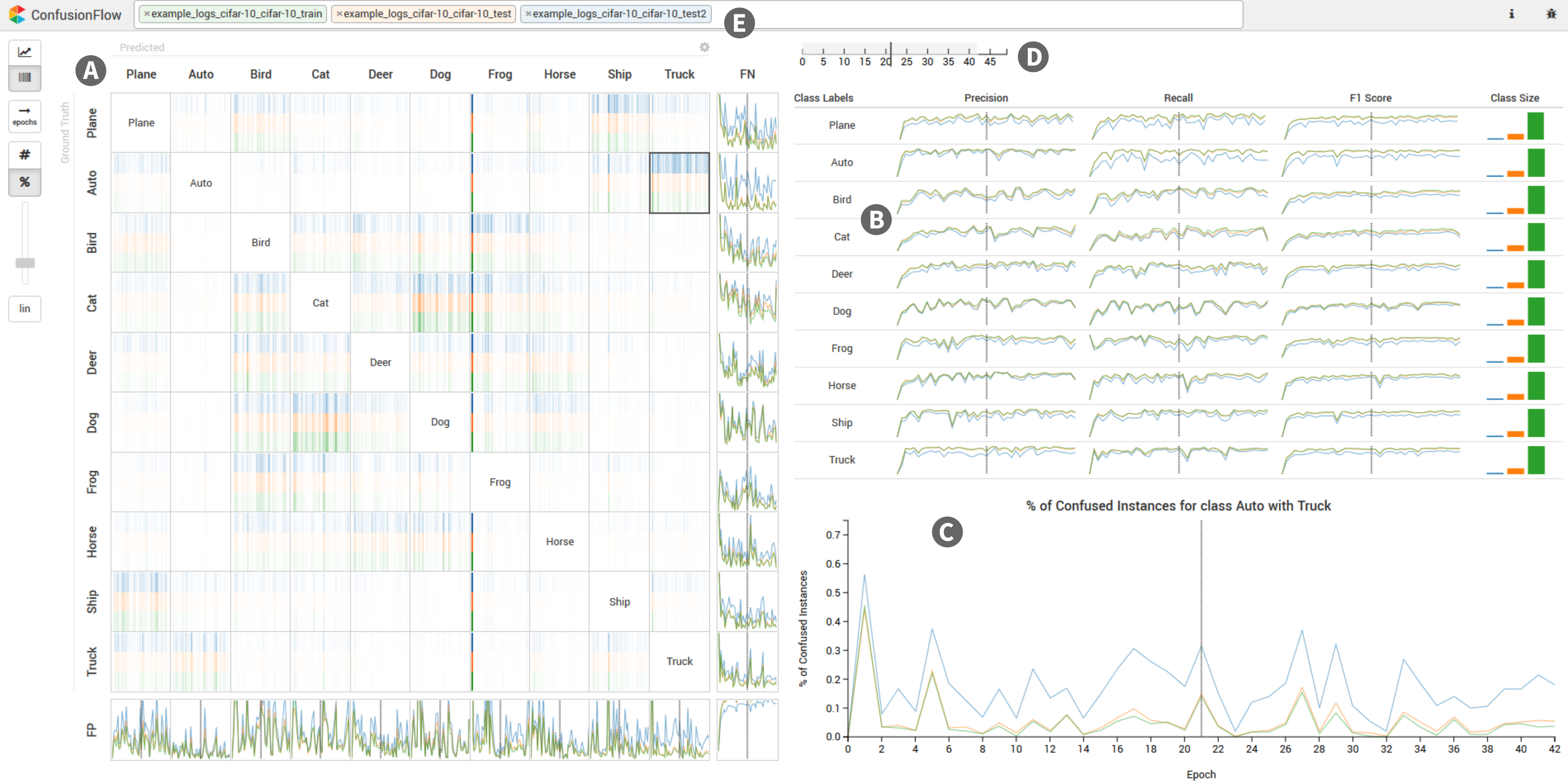}
    \caption{%
    The ConfusionFlow matrix~(\protect\annotref*{a}) visualizes confusion of classifiers across training iterations.
    Performance data for multiple classifiers can be loaded~(\protect\annotref*{e}) and compared with each other.
    Additionally, class-wise performance measures and class distributions are displayed in a second view~(\protect\annotref*{b}).
    The timeline~(\protect\annotref*{d}) allows interactive exploration and selection of temporal regions of interest.
    On demand, plots can be expanded to the detail view~(\protect\annotref*{c}).
    Here, we compare the performance of a neural network classifying images from the train set (\protect\colorswatch{CFgreen}) and test set (\protect\colorswatch{CForange}) of CIFAR-10~\cite{krizhevsky_learning_2009}, and a recently proposed \CFreplace{new}{alternative} test set (\protect\colorswatch{CFblue}) from CIFAR-10.1~\cite{recht_cifar-10_2018}, respectively.
    The line chart (C) shows that the relative number of misclassified images for the selected classes \emph{auto} and \emph{truck} deviates notably between the original and the new test set.
    For the remaining classes the classifier performs similarly on the new test set and the original CIFAR-10 test set.
  }
  \label{fig:teaser}
\end{figure*}

The ConfusionFlow interface consists of three views, as illustrated in Figure~\ref{fig:teaser}:
\CFinsert{(\annotref*{a})~}the \emph{ConfusionFlow matrix} presenting the confusion of one or more classifier(s) over time\CFdelete{~(\annotref*{a})};
\CFinsert{(\annotref*{b})~}the \emph{class performance and distribution view}, including plots of precision, recall, and \(F_1\)-score, as well as visualizations of the instances' class distributions\CFdelete{~(\annotref*{b})};
and \CFinsert{(\annotref*{c})~}the \emph{detail view} showing magnified plots of interactively selected confusion or performance curves\CFdelete{~(\annotref*{c})}.
Additionally, ConfusionFlow features \CFinsert{(\annotref*{d})~}a \emph{timeline} for selecting the range of training steps that are used for exploration\CFdelete{~(\annotref*{d})};
and \CFinsert{(\annotref*{e})~}an input field for loading datasets\CFdelete{~(\annotref*{e})}, which also serves as a legend for the whole visualization.

Figure~\ref{fig:teaser} shows how ConfusionFlow can be used to compare the image classification performance of a neural network on different datasets.
For this example, we have loaded confusion data for a neural network image classifier trained on the training set (\colorswatch{CFgreen}) of  CIFAR-10~\cite{krizhevsky_learning_2009}, and evaluated on the images from the corresponding test set (\colorswatch{CForange}), as well as on a recently proposed new test set (\colorswatch{CFblue}) from CIFAR-10.1~\cite{recht_cifar-10_2018}, respectively.

\subsection{ConfusionFlow Matrix}
\label{sec:confusion-matrix}

\begin{figure}
    \centering
    \includegraphics[width=\columnwidth]{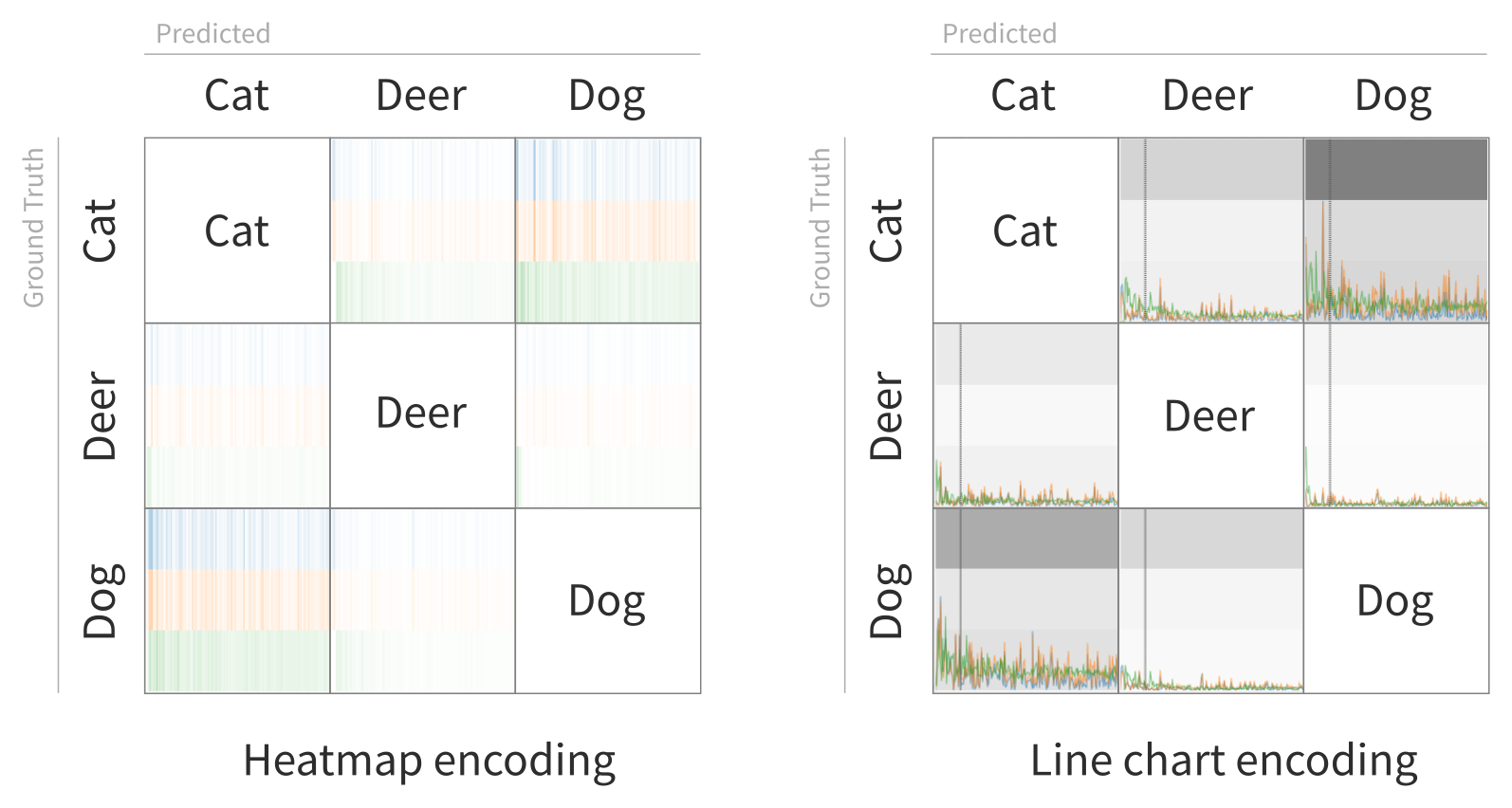}
    \caption{The ConfusionFlow matrix with its two different encoding options.
    Left: Stacked heatmap encoding (lasagna plot~\cite{swihart_lasagna_2010}).
    Right: Superimposed line charts with background heatmap corresponding to selected iteration.}
    \label{fig:ui-vis-matrix}
\end{figure}

The ConfusionFlow matrix, shown in Figures~\ref{fig:teaser}\annotref{a} and \ref{fig:ui-vis-matrix}, is a visualization of classification errors that supports the within- and between model  comparison (\ref{hl-task:comp}) as well as temporal analysis (\ref{hl-task:temp}).
In the classic confusion matrix, cell \((i,j)\) lists the number of instances with ground truth label of class \(i\) that are classified as class \(j\).
While the classic confusion matrix is limited to showing confusion data for a single model at one specific time step, the ConfusionFlow matrix visualizes the class confusion for \emph{multiple classifiers} over \emph{multiple training steps} (see Figure~\ref{fig:perf-levels}).
As described in Section~\ref{sec:comparison-and-temporal}, the different classifiers can come from ML experiments with different models or ML experiments with different datasets or dataset folds.
The ConfusionFlow matrix is a small multiples approach: for each cell, the single value of the classic confusion matrix is replaced with a plot of the values for a selected time range (see Figure~\ref{fig:ui-vis-matrix}).

\begin{wrapfigure}[8]{r}{0pt}%
    \includegraphics[height=6\baselineskip]{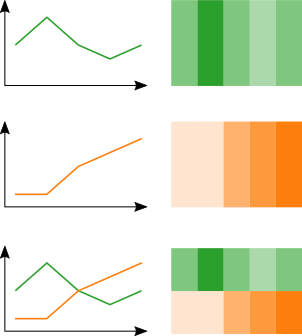}%
\end{wrapfigure}
The ConfusionFlow matrix should enable temporal analysis~(\ref{hl-task:temp}) and comparison~(\ref{hl-task:comp-between}) at the same time, while conserving the familiar layout of the confusion matrix.
This means that the confusion visualization for each classification model should only take up little space, but should still be able to show a fine-grained temporal resolution.
At the same time, individual temporal progressions for different models should be easily distinguishable to enable users to perform tasks \ref{task:measure-prog} and \ref{task:comp-prog}.
Accordingly, we chose the heatmap idiom for the visualization of class confusions \cite{bernard_visualinteractive_2019}.
One-dimensional heatmaps, sometimes called colorfields, have been shown to support the task of time series comparison well, particularly in terms of task completion time~\cite{gogolou_comparing_2019}.
The thumbnail at the top right of this paragraph shows how temporal confusion values for a single cell of the matrix are encoded in this idiom.
Each matrix cell \CFreplace{has}{contains} one time series per loaded model.
In a line chart encoding, multiple time series are plotted together in the same chart \CFinsert{(superposition strategy, cf.~Gleicher et al.~\cite{gleicher_visual_2011})}.
If each line is instead transferred to a one-dimensional heatmap, the heatmaps can be stacked for comparison without any overplotting issues \CFinsert{(juxtaposition strategy~\cite{gleicher_visual_2011})}.
These stacked heatmaps are sometimes called \enquote{lasagna plots}~\cite{swihart_lasagna_2010}.
The confusion value is encoded as brightness,
and a unique hue is automatically assigned to each classifier.
To ensure visual consistency, \CFreplace{hues of}{the hue for each} classifier\CFdelete{s} \CFreplace{are}{is} kept constant across all linked views.
The left part of Figure~\ref{fig:ui-vis-matrix} shows a matrix made up of such stacked heatmaps in a real example.

\begin{wrapfigure}[16]{r}{0pt}%
    \includegraphics[height=14\baselineskip]{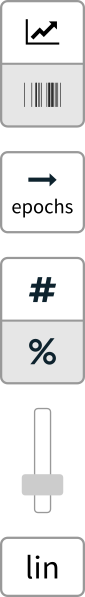}%
\end{wrapfigure}
Users can interactively switch from the heatmap to a line chart encoding of class confusion.
This option is available for two reasons.
First, we found that ML users are particularly used to viewing temporal performance measures as line charts.
Second, the line charts can facilitate reading off and comparing absolute values, either for individual epochs of interest (\ref{task:measure-perf}) or between models (\ref{task:comp-perf}).
\CFdelete{After switching to the line chart encoding, plots of multiple classifiers are shown as superimposed lines.}
If the line chart encoding is selected and the user additionally selects a single iteration (see Section~\ref{sec:timeline} for information on the timeline  selector), then a stacked heatmap of the confusion values for the selected iteration is plotted as background for the line charts.
An example of this encoding is shown in the right part of Figure~\ref{fig:ui-vis-matrix}.
The gray background heatmaps correspond to the confusion values for the three models at the time step indicated by the dashed vertical line.
This additional background heatmap increases the visual saliency of problematic class-pairs for the line chart encoding
This increased saliency is already inherent in the heatmap encoding.
The heatmap\slash line chart switch, as wells as several other controls for visual encoding options, can be seen to the left of the ConfusionFlow matrix in Figure~\ref{fig:teaser}.

To facilitate comparison of cells for a given predicted class, time by default progresses left-to-right, regardless of the encoding choice.
This choice lines up with most users' expectations for plots of time series data.
On demand, users can rotate the cell contents by 90\(^\circ\) for easier comparison along a given ground truth class, if the heatmap encoding is selected.

The diagonal elements of the classic confusion matrix list the numbers of correctly classified instances for each class.
For well-functioning classifiers, these numbers are typically much higher than the confusion counts.
To keep the user's focus on the exploration of the error behavior and to keep lookup of confusion values feasible (\ref{task:measure-perf} and \ref{task:comp-perf}), we decided to replace the diagonal cells in the ConfusionFlow matrix by class labels.
In this way, we retain high visual contrast in the off-diagonal cells, and facilitate navigating through the matrix.

To let users assess the overall performance of individual classes, we show temporal plots of false negatives for each class in an additional column to the right, slightly offset from the matrix.
Likewise, an additional row at the bottom shows the false positives for each class.
We use the diagonal element at the intersection of the additional row and column to show the temporal progression of the overall accuracy of the classifier(s).
This allows the user to perform all analysis tasks \ref{task:measure-perf} to~\ref{task:relate-problem} on a global level (\ref{lvl:global})---especially when this chart is brought to the detail view (see Section~\ref{sec:detail-view}).

To enable performance comparison between datasets of different sizes (\ref{task:comp-perf} to~\ref{task:relate-problem}), such as training versus test sets, ConfusionFlow includes an option to switch from absolute to relative performance values.
To obtain the relative performance scores, the confusion counts are simply divided by the total number of classified instances.

In order to address anomaly detection and comparison tasks (\ref{task:detect-problem} and~\ref{task:relate-problem}), peak values for problematic pairs of classes or training iterations should be visible and salient by default.
However, these particularly high confusion counts can sometimes hide potentially interesting findings in other cells.
To address this issue, we let users toggle between linear and logarithmic scales for the plots.
Using an exponential scaling slider, lower values can be further accentuated\CFreplace{, which essentially adjusts the contrast in case of the heatmap visualization.}{. In the heatmap encoding, this corresponds to increasing the contrast.}

If users are only interested in a subset of the class alphabet, they can narrow down the number of selected classes in a \emph{class selection} dialog.
In order to maintain a meaningful confusion matrix, a minimum of two classes need to be selected at all times.
The number of displayed classes is not limited in terms of implementation, but there are some \CFdelete{clear} practical limitations which we discuss in depth in Section~\ref{sec:eval-scalability}.
\CFinsert{\emph{Class aggregation} is a second strategy to reduce the number of displayed classes.
This strategy is currently not supported within ConfusionFlow and is only relevant for datasets with a distinct class\slash superclass hierarchy.}

In case of the CIFAR-10 example shown in Figure~\ref{fig:teaser}, the ConfusionFlow matrix reveals that the confusion between classes \emph{auto} and \emph{truck} is considerably higher for the CIFAR-10.1 test set.
Due to ConfusionFlow's focus on temporal analysis, it is immediately visible that this error is consistent across all training epochs (cf.~tasks \ref{task:comp-perf} and~\ref{task:comp-prog}).
For all other pairs of classes, the network generalizes well\CFdelete{ to the new dataset}.
In those cells, this can be seen by the similar brightness values for all three selected datasets.
Without these class-level observations, the reason for the decrease in overall accuracy would remain unclear; by performing a comparative analysis with ConfusionFlow, the performance decrease \CFreplace{could}{can} be traced back to changes in the underlying data distribution.

\subsection{Class Performance \& Distribution View}
\label{sec:performance-measures}

A thorough, class-level (\ref{lvl:class}) analysis of a classifier's performance should not only focus on pairs of classes, but also include assessment of the general performance for \emph{individual} classes.

\begin{wrapfigure}[6]{r}{0pt}%
    \includegraphics[width=0.23\linewidth]{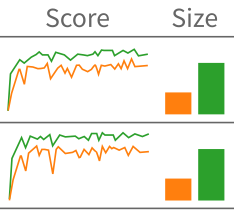}%
\end{wrapfigure}
To this end, ConfusionFlow provides temporal (\ref{hl-task:temp}) line charts of precision, recall, and \(F_1\)-score for all selected classes (see Figure~\ref{fig:teaser}\annotref{b}).
In contrast to the ConfusionFlow matrix, horizontal space is not as limited for these plots, and visual clutter is thus less of an issue even in case of noisy data.
For comparison between multiple classifiers (\ref{hl-task:comp-between}), we superimpose the line charts.
Again hue is used to differentiate \CFreplace{multiple}{between different} classifier results\CFdelete{ visually}, and the hues are consistent with those chosen for the ConfusionFlow matrix.
To let users assess the size and class distributions for each dataset, bar charts encode the number of instances per class.
The bars are placed next to the per-class performance metrics and are\CFdelete{ again} colored consistently with all other views.
A~mouseover action reveals the number of instances for each bar.

In case of the CIFAR example (Figure~\ref{fig:teaser}), the 
class distribution charts reveal that the updated test set from CIFAR-10.1 is considerably smaller than that from CIFAR-10.
The precision, recall, and \(F_1\)-score charts confirm that the recall for class \emph{auto} and the precision for class \emph{truck} are particularly bad for the new test set, but they also suggest that the classifier could not generalize well for \emph{plane} instances from CIFAR-10.1.
This assessment of generalization capabilities over multiple epochs is an example of task \ref{task:comp-prog}.
While the class performance charts can lead to interesting insights on their own, they should also lead the user to re-examining the ConfusionFlow matrix.

\subsection{Detail View}
\label{sec:detail-view}

All cells in the ConfusionFlow matrix, as well as all per-class performance charts can be selected to be shown in greater detail in a separate view (see Figure~\ref{fig:teaser}\annotref{c}).
We visualize the temporal development of selected cells as line charts and superimpose the curves for multiple classifiers, keeping the hue for each model\slash configuration consistent.
The detail view particularly addresses the tasks of pin-pointed temporal identification of problematic instances (\ref{task:detect-problem} and \ref{task:relate-problem}) as well as reading off and comparing numeric values (\ref{task:measure-perf} and \ref{task:comp-perf}), as space in the ConfusionFlow matrix is rather limited.
Upon loading a new performance dataset, by default the overall (global) accuracy is shown in the detail view, as users are accustomed to this plot from many other performance analysis tools.

For the CIFAR example, the detail view confirms that---for a typical iteration---the confusion value of interest (\emph{auto} vs.~\emph{truck}) is about twice as high for the updated CIFAR-10.1 test set than for the CIFAR-10 test and train sets.

\subsection{Timeline}
\label{sec:timeline}

\begin{wrapfigure}[2]{r}{.5\columnwidth}%
    \vspace*{-1ex}%
    \includegraphics[width=.49\columnwidth]{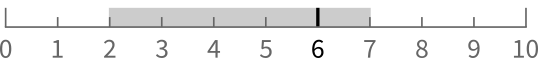}%
    \vspace*{1ex}%
\end{wrapfigure}
ConfusionFlow should aid users in exploring the error behaviour of classifiers at different temporal granularities, involving single-time step tasks (\ref{task:measure-perf} and \ref{task:comp-perf}) and multiple-time-step tasks (\ref{task:measure-prog}, \ref{task:detect-problem}, \ref{task:comp-prog}, and \ref{task:relate-problem}).
Moving from a temporally rough to a more fine-grained analysis is facilitated by the timeline shown in Figure~\ref{fig:teaser}\annotref{d}.

By default, the whole range of available iterations is selected upon loading performance data.
Users can select a subset of iterations by dragging the left and right boundaries of the highlighted range in the timeline.
All linked views, such as the ConfusionFlow matrix or the detail view, are updated automatically according to the selection.
This range selection corresponds to a temporal zoom-in.

To support users in locating and comparing values for a specific time step across multiple views, they can additionally select a single training step by clicking the iteration number below the range selector.
A black line in the timeline marks the currently selected single iteration.
The selected iteration is then dynamically highlighted by a vertical marker in all linked components.
If the line chart encoding is selected for the ConfusionFlow matrix, the background heatmap will also be updated as described in Section~\ref{sec:confusion-matrix}.

In the CIFAR example from Figure~\ref{fig:teaser}, the performance data spans 50~epochs.
The user has selected an epoch range covering epochs~0 to~42, and found an interesting peak for the confusion \emph{auto} vs.~\emph{truck} at epoch~22 in the detail view.


\subsection{Dataset Selection}

\begin{wrapfigure}[2]{r}{.5\columnwidth}%
    \vspace*{-2ex}%
    \includegraphics[width=.49\columnwidth]{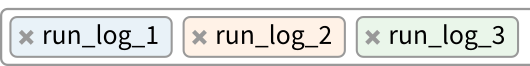}%
    \vspace*{2ex}
\end{wrapfigure}
As \CFreplace{mentioned}{stated} above, a unique hue is automatically assigned to each classification run upon loading the respective performance data.
An input field with dynamic drop-down suggestions lets the user select from pre-loaded performance data for a number of classification configurations (see Figure~\ref{fig:teaser}\annotref{e}).
After the user made the selection, the input field serves as a legend for the visualization, representing each run with a colored box.

ConfusionFlow is a model-agnostic visualization technique. This means that the kind of data on which ConfusionFlow relies does not depend on the nature of the model.
During training, only the classifier output for each instance needs to be logged after each iteration, and stored along with the ground truth labels.
Along with our prototype implementation (see Section~\ref{sec:implementation} below) we provide Python code examples for logging and exporting data for the commonly used ML frameworks TensorFlow and PyTorch.

In Figure~\ref{fig:teaser}, the input field serves as a legend to remind the user that performance data for the training (\colorswatch{CFgreen}) and test set (\colorswatch{CForange}) of CIFAR-10~\cite{krizhevsky_learning_2009}, as well as the recently proposed new test set (\colorswatch{CFblue}) from CIFAR-10.1~\cite{recht_cifar-10_2018}, have been loaded.

\subsection{Implementation}
\label{sec:implementation}

ConfusionFlow is a server-client application based on the Caleydo Phovea framework\footnote{Caleydo Phovea: \href{https://github.com/phovea/}{https://github.com/phovea}} and the Flask framework\footnote{Flask: \href{http://flask.pocoo.org/}{http://flask.pocoo.org}}.
The server side is written in Python and the client side is written in TypeScript using D3.js~\cite{bostock_d3:_2011}.
The code for ConfusionFlow---including the logging tools mentioned above---is available on GitHub\footnote{ \href{https://github.com/ConfusionFlow/confusionflow}{https://github.com/ConfusionFlow/confusionflow}}.

A~deployed prototype of ConfusionFlow with \CFreplace{some}{several} pre-loaded example datasets is available \CFreplace{at \href{https://confusionflow.caleydoapp.org/}{https://confusionflow.caleydoapp.org}}{online}\CFinsert{\footnote{Prototype: \href{https://confusionflow.caleydoapp.org/}{https://confusionflow.caleydoapp.org}}}.

\section{Evaluation}

To evaluate the usefulness of ConfusionFlow we describe \CFreplace{a case study performed}{three scenarios. First, we describe a case study which we performed} in collaboration with ML researchers.
Our collaborators used ConfusionFlow for visually comparing labelling strategies in active learning.
In a second evaluation scenario, we \CFreplace{look at possible scalability issues of when applied}{analyze how} ConfusionFlow\CFdelete{, when applied} \CFinsert{scales} to datasets with many \CFinsert{(\(\gtrsim 10\))} classes\CFdelete{, and we compare two different strategies to work around these issues}.
\CFinsert{A~third use case shows how ConfusionFlow can be applied to study the effects of different neural network pruning strategies.}

\subsection{Case Study: Effective Labeling in Active Learning}
\label{sec:case-study-active-learning}

\begin{figure*}[t]
    \centering
    \includegraphics[width=1.0\linewidth]{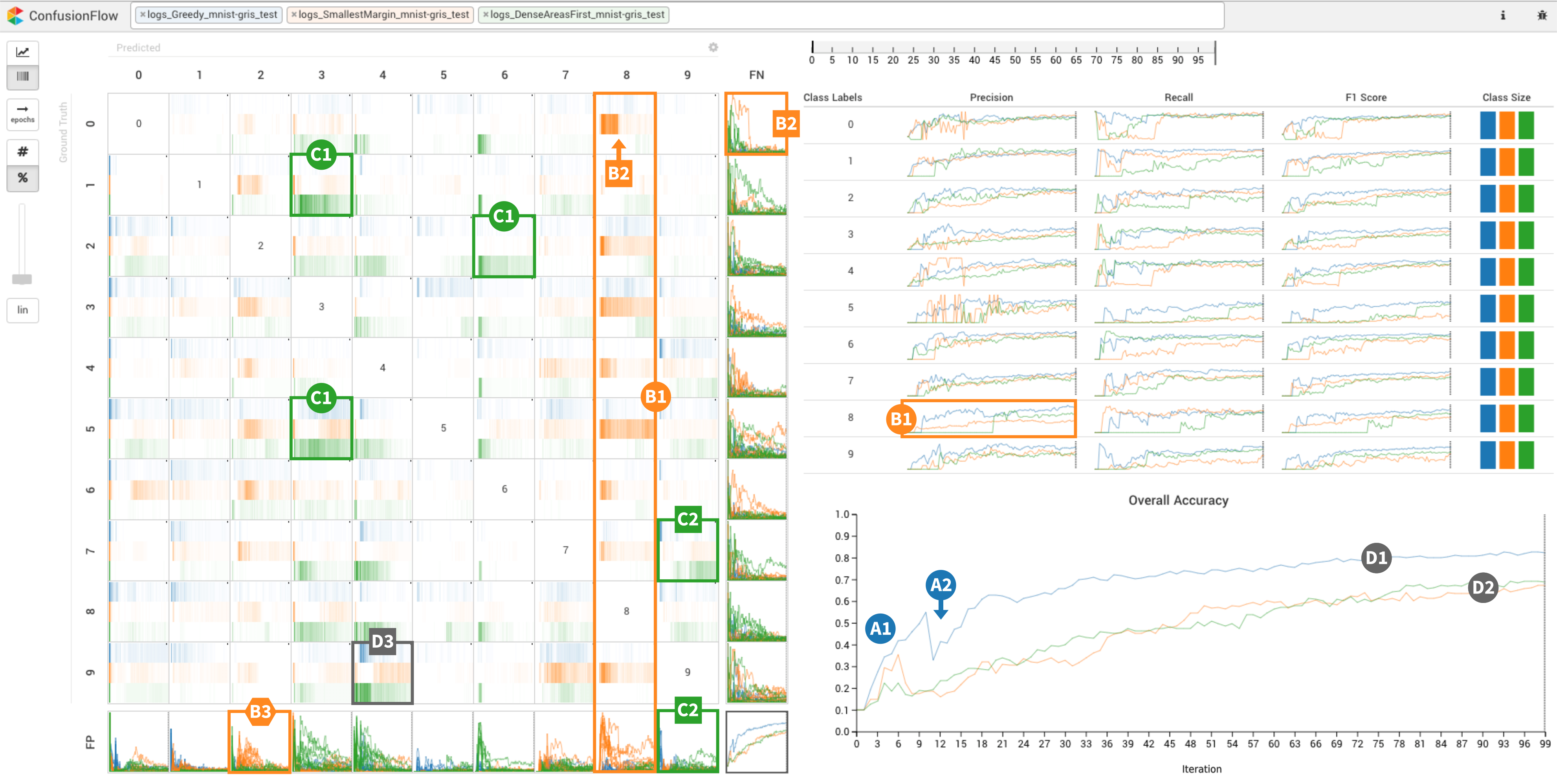}
    \caption{%
    Visual comparison of instance selection strategies for effective data labeling.
    In an experiment, our collaborators tested three different labeling strategies: Greedy~(\protect\colorswatch{CFblue}), Smallest Margin~(\protect\colorswatch{CForange}), and Dense Areas First~(\protect\colorswatch{CFgreen}).
    Using ConfusionFlow, our collaborators made a series of findings regarding the overall performances~(\protect\annotref*{a, d1, d2}) as well as the temporal progression of class confusions~(\protect\annotref*{b, c, d3}) for the different strategies.}
\label{fig:labelingUseCaseResult}
\end{figure*}

\CFreplace{For virtually any supervised ML task, labeled (ground truth) data for training and testing ML models is an essential prerequisite.}{Labeled data is a prerequisite for supervised ML tasks.}
The \CFdelete{iterative} labeling process requires human supervision\CFdelete{, as humans need} to attach \CFreplace{\enquote{semantic} information}{semantics} to data instances.
\CFdelete{Subsequently, labels and associated instances can be used as training data.}
The challenge addressed in this case study refers to the instance selection problem: which instance should be selected \CFinsert{next} for labeling\CFdelete{ next}, in order to
\begin{enumerate*}[
    label={(\alph*)},
    itemjoin={{, }},
    itemjoin*={{, and }},
    after={?}]
\item improve a classifier most effectively
\item keep the effort of human labelers \CFdelete{(oracles) }at a minimum
\end{enumerate*}

\CFdelete{In the ML field, active learning~\cite{settles_active_2012} is a class of model-driven approaches that aim at supporting the meaningful selection of instances in this incremental labeling process, while humans only label these selected instances.}
\CFinsert{Within the field of active learning~\cite{settles_active_2012}}, \emph{Visual-Interactive Labeling} (VIAL)~\cite{bernard_vial:_2018} is a concept to combine the strengths of humans and algorithmic models to facilitate effective instance selection.
To this end, VIAL combines active learning with visual-interactive interfaces which enable humans to explore and select instances~\cite{bernard_comparing_2018}.

\CFreplace{The}{One model-based} active learning strategy used in this study is \emph{Smallest Margin}~\cite{wu_sampling_2006}, \CFinsert{which} always select\CFreplace{ing}{s the remaining unlabeled} instances with \CFinsert{the} highest classifier uncertainty.
\CFreplace{O}{In contrast, o}ne instance selection strategy frequently applied by humans is \emph{Dense Areas First}, reflecting the idea that humans tend to select instances in the dense areas of the data~\cite{bernard_towards_2018}.

Our collaborators aim at making the labeling process a more effective, efficient, and human-friendly endeavor. 
Given that recent experiments confirmed that human-based strategies and model-centered strategies have complementary strengths~\cite{bernard_towards_2018}, \CFreplace{it will be interesting to}{our collaborators are interested in} further \CFreplace{explain}{analyzing the} differences \CFreplace{across}{between} strategies.

ConfusionFlow allows our collaborators to compare the model-based \emph{Smallest Margin} with the human-based \emph{Dense Areas First} strategy.
\CFinsert{As a third strategy, a \emph{Greedy} algorithm based on ground truth information serves as the theoretical upper-limit of performance.}
For their analyses, our collaborators chose the MNIST handwritten digits dataset~\cite{lecun_gradient-based_1998}\CFdelete{,} as an intuitive and well-established dataset that does not require domain knowledge.
Another advantage of MNIST over other datasets is the ability of users to label most instances unambiguously.

\CFdelete{\subsubsection*{Visual Comparison of Instance Selection Strategies}}

Our collaborators use ConfusionFlow's temporal analysis capability to analyze and compare the labeling process over time.
Accordingly, each training epoch corresponds to one labeling iteration (consisting of instances selection, labeling, and model re-training).
Figure~\ref{fig:labelingUseCaseResult} shows how the collaborators compared the labeling processes of the three strategies visually (Smallest Margin, Dense Areas First, and Greedy).

The primary comparison goal (\ref{hl-task:comp}) of our collaborators is between\CFdelete{-} models, accompanied by more detailed analyses of individual within-model characteristics.
Due to the exploratory nature of the \CFreplace{use case}{analysis}, all three temporal analysis goals (\ref{hl-task:temp}) are relevant.
As a result, the information requirements of our collaborators include all six analysis tasks (\ref{task:measure-perf} to~\ref{task:relate-problem}, see Table~\ref{tab:tasks}).

The Greedy strategy (Figure~\ref{fig:labelingUseCaseResult}, \colorswatch{CFblue}) shows a steep performance increase (\ref{task:measure-prog}) at the beginning~(\annotref*{a1}), leading to more than 50~percent accuracy after only 10 iterations (\ref{task:measure-perf}).
\CFreplace{Compared to the other strategies}{As the theoretical upper performance limit}, Greedy has the strongest upward trend \CFreplace{and}{compared to the other strategies. It} converges earlier (\ref{task:comp-prog}) \CFreplace{,}{and} at the highest level of accuracy (\ref{task:comp-perf}).
\CFdelete{This behavior was expected as this strategy uses ground truth information to depict the theoretical upper-limit of performance.}
With only 50 labels the Greedy strategy already achieves almost 80~percent accuracy\CFinsert{~(\ref{task:measure-perf})}.

Our collaborators also identified a known anomaly pattern in the accuracy curve (\ref{task:detect-problem}), which happens after ten instances (i.e., when all labels have been visited exactly once; see Figure~\ref{fig:labelingUseCaseResult}\annotref{a2}).
This pattern is unique \CFreplace{in}{to} the Greedy strategy~(\ref{task:relate-problem}): with the eleventh label the training set becomes unbalanced, leading to a significant temporal decrease of classification accuracy.
With ConfusionFlow, our collaborators \CFreplace{were able to analyze}{could relate} this anomaly \CFreplace{at a more fine-grained level.
The matrix view reveals that the anomaly pattern can be explained by}{to increased} confusions between the classes~0, 4, and 9~(\ref{task:comp-perf}).
Future repetitions of the experiment will clarify whether this effect can be related to the semantics of the particular classes or can be explained by other influencing factors.

The Smallest Margin strategy (Figure~\ref{fig:labelingUseCaseResult}, \colorswatch{CForange}) starts with a short and very early peak (instances~3 to~6)~(\ref{task:detect-problem}) a pattern that the other two strategies do not show~(\ref{task:relate-problem}).
Thereafter, an almost linear upwards trend continues until instance~50~(\ref{task:measure-prog}), where the Margin strategy has almost 60~percent accuracy~(\ref{task:measure-perf}).

In the ConfusionFlow matrix, our collaborators identified considerably high confusion values of class~8 (\ref{task:measure-perf}) with almost any remaining class (Figure~\ref{fig:labelingUseCaseResult}\annotref{b1}).
This poor performance for class~8 is also clearly visible in the precision curve.
An interesting pattern was the significant decrease of confusion between classes~0 vs.~8, roughly beginning at the 35th time step~(\annotref*{b2}).
It seems that sometimes a single labeled instance can make a difference and support the accuracy of a classifier.
Additionally, up to around instance~50, confusion values for class~2 are relatively high (\ref{task:detect-problem}), leading to many false positives for this class (\annotref*{b3}).

The Dense Areas First strategy (Figure~\ref{fig:labelingUseCaseResult}, \colorswatch{CFgreen}) exhibits a minor but steady increase in the early phase of the labeling process (\ref{task:measure-prog}).
After 50 labels, the strategy achieves almost 55~percent accuracy \ref{task:measure-perf}.
At a glance, Dense Areas First and Smallest Margin have similar overall accuracy curves (\ref{task:comp-prog}).

By inspecting the ConfusionFlow matrix, the analysts gained a series of different insights.
Some class confusions lasted for the entire labeling process (\ref{task:measure-prog}) (1~vs.~3, 5~vs.~3, 2~vs.~6; see Figure~\ref{fig:labelingUseCaseResult}\annotref{c1}). 
Conversely, some class confusions seemed to have different levels of intensity during the labeling process (\ref{task:measure-prog}) (2~vs.~4, 7~vs.~4, 9~vs.~4).
One class confusion even increased during the process (\ref{task:measure-prog}) (7~vs.~9), visible both in the matrix and in the FP chart~(\annotref*{c2}).
Some training steps introduced peaks of confusion (\ref{task:detect-problem}) (class 6, roughly at instance 10).
Finally, some classes did not suffer from considerable confusions at all (0, 1, 5, and 7).

\CFinsert{One of the most interesting findings---according to our collaborators---was that the confusion patters for some pairs of classes differed considerably over time and between strategies.
In case of the classes 9~vs.~4, for example, the confusions of the model-based Smallest Margin and the human-based Dense Areas First strategy show a strongly contrasting behavior (see Figure~\ref{fig:labelingUseCaseResult}\annotref{d3}).
This observation strengthens our collaborators' view that strategies have complementary strengths.

As a result of their analysis with ConfusionFlow, our collaborators are motivated to perform in-depth follow-up analyses.
They are particularly interested in using their class-level insights~(\ref{lvl:class}) gained with ConfusionFlow as a starting point for drilling down to the instance level~(\ref{lvl:instance}).
This way, they will be able to confirm general class-related patterns or trace back performance changes to individual images.
}

\CFdelete{
\subsubsection*{Summary of Findings}

Being able to compare model performances with the overall accuracy visualization helped our collaborators to start their analysis in a way familiar to ML experts.
Building on familiar concepts, the detailed analysis of individual characteristics of training processes was new to the collaborators.
Both the ConfusionFlow matrix and the Class Performance \& Distribution View offered new perspectives on labeling processes. 
At a glance, our collaborators identified a transition from \emph{detecting} known\slash unknown patterns to \emph{explaining} pattern characteristics using the more detailed analysis capability.

Along these lines and according to the analysts, one of the most interesting findings at the granularity of class confusions was that for some pairs of classes the class confusion patterns differed considerably over time.
In case of the classes 9~vs.~4, for example, the confusions of the model-based Smallest Margin and the human-based Dense Areas First strategy showed strongly contrasting behavior (see Figure~\ref{fig:labelingUseCaseResult}\annotref{d3}).
The general observation that class confusions over time can be vary strongly for different strategies indicates that these strategies have complementary strengths.
This finding encouraged our collaborators to further investigate such effects at the granularity of classes and class confusions.
These differences also became apparent from the FP and FN plots, which our collaborators did not expect to differ that strongly.

In summary, the findings lead to interesting follow-up questions such as:
\begin{enumerate*}[
    label={(\alph*)},
    itemjoin={{; }},
    itemjoin*={{; and }},
    after={?}]
\item how do the identified temporal effects relate to the data characteristics?
\item how do temporal performance trends add to the complementing strengths of selection strategies?
\item how can class-level characteristics be used to create better instance selection strategies in the future
\end{enumerate*}
Finally, our collaborators mentioned that they would be particularly interested in using their findings from ConfusionFlow at a class-based level as a starting point for drilling further down to the instance level.
}

\CFdelete{%
The pruning usage scenario and the case study on labeling strategies in active learning both highlighted the importance of being able to perform model comparison~([...]) and temporal analysis of learning~([...]) \emph{at the same time}.
In both contexts ConfusionFlow allowed the users to trace back changes in the overall accuracy to the confusion counts for individual pairs of classes~([...]).
This way, our visualization approach can help answer questions such as whether the model errors change uniformly over all classes, whether they are limited to certain difficult pairs, or how these errors change between training iterations.
It also became obvious that ConfusionFlow should not be the last step in the workflow for analyzing classifier performance.
Our collaborators mentioned that they would be particularly interested in using their findings from ConfusionFlow as a starting point for drilling further down to the instance level.
We also gained important feedback about some other limitations of ConfusionFlow, which we will discuss in the following section.%
}

\CFinsert{\subsection{Use Case: Scalability to Many Classes}}
\label{sec:eval-scalability}

\CFreplace{One of the limitations of the ConfusionFlow matrix visualization is its scalability.
Since t}{T}he \CFinsert{traditional} confusion matrix\CFreplace{ is }{---as} a\CFreplace{n}{~class-level} aggregation technique\CFreplace{, it }{---}scales well to large datasets with many instances.
However, the confusion matrix \CFdelete{is a way to} organize\CFinsert{s} the inter-class confusion counts across \emph{all pairs of dataset classes}, so datasets with more classes result in larger confusion matrices.
Doubling the number of classes in a dataset reduces the available area for each cell in the confusion matrix by a factor of four.

\CFinsert{The ConfusionFlow visualization idiom inherits these scalability issues from the traditional confusion matrix}.
In its current implementation, ConfusionFlow works well for classification problems with up to around fifteen classes.
This is sufficient to support the majority of freely available datasets \CFinsert{for multiclass classification}\footnote{\CFinsert{In principle, ConfusionFlow can also be applied to binary classification tasks, but the analysis of binary classifiers does not benefit much from ConfusionFlow's focus on the class level~(\ref{lvl:class}).}}: a query on \url{www.openml.org}~\cite{vanschoren_openml_2014} in February~2020 reveals that out of \CFreplace{1,302}{434} datasets for \CFinsert{multiclass} classification problems, \CFreplace{1,236}{368} datasets have fewer than 15~classes.


\CFdelete{%
Still, we would like to assess the specific problems that might arise when attempting to use ConfusionFlow for classification problems with a high number of classes, and how to possibly work around any limitations.
Since the ConfusionFlow matrix encoding requires a certain minimum amount of space, the only way to use it in cases of many classes is to reduce the number of cells shown.
For this reduction, two basic techniques exist: class \emph{selection} and class \emph{aggregation}.
In class selection, only a subset of the classes are actually used for constructing the confusion matrix, and all other classes are disregarded.
In class aggregation, so-called superclasses are constructed.
Each superclass comprises multiple classes that are considered to be similar in some sense.}

\CFinsert{Still, given the popularity of certain benchmark datasets for image classification with higher numbers of classes---such as ImageNet~\cite{deng-imagenet-2009} (1000 classes), Caltech 101~\cite{feifei-learning-2004} and 256~\cite{griffin-caltech-2007} (101 and 256 classes, resp.), and CIFAR-100~\cite{krizhevsky_learning_2009} (100 classes)---it is worth evaluating ConfusionFlow's scalability.

To this end, we assess the characteristics of the ConfusionFlow matrix in the context of two class reduction strategies, class \emph{selection} and class \emph{aggregation} (cf.~Section~\ref{sec:confusion-matrix}), when applied to an image classifier trained on the CIFAR-100 dataset.}

\CFreplace{In order to investigate how well these two different mitigation mechanisms for scalability issues---class selection and class aggregation---perform, we apply ConfusionFlow to a classification task on the CIFAR-100 dataset~\cite{krizhevsky_learning_2009}.
In contrast to CIFAR-10, which features only ten classes, CIFAR-100 features one hundred classes.
These one}{The one} hundred~classes \CFinsert{of CIFAR-100} are semantically divided into twenty superclasses.
Each superclass groups five classes, e.g., the classes \emph{fox}, \emph{porcupine}, \emph{possum}, \emph{raccoon}, and \emph{skunk} together make up the superclass \emph{medium-sized mammals}.
We chose this dataset as it already includes a proposed class aggregation scheme, making it suitable for an unbiased comparison between the two approaches for reducing the dimensions of the confusion matrix.


We trained a simple convolutional neural network to classify images from the CIFAR-100 dataset into one of the 100~classes, logging confusion data for 100~epochs.
After each epoch, we evaluated the network's performance on the train fold as well as the test fold of the whole dataset.
\CFreplace{Additionally to the class confusion values for the 100~classes, we recorded confusion values for the 20~superclasses.}{We determined superclass confusion values from the class predictions.}
\CFdelete{We did not focus on network architecture or hyperparameter tuning and thus did not achieve state-of-the-art classification accuracy.
However, we believe that our classifier may be representative of the \enquote{not so well-working} classifiers that a typical user would have to evaluate as part of the development of production-ready ones.}

We \CFreplace{first }studied \CFdelete{different strategies for }class selection \CFinsert{based on two different selection criteria: \(F_1\) scores and off-diagonal confusion matrix entries.}
\CFdelete{The insights gained by ConfusionFlow matrices from a subset of the classes depends highly on the criterion chosen for selecting the classes.}
We first selected the ten classes with the lowest \(F_1\)-scores on the test set in the last epoch.
\CFdelete{The \(F_1\) evaluation was carried out in Python prior to the analysis with ConfusionFlow, but data-dependent class selection could be added to future versions of ConfusionFlow, as the necessary data is already included in the confusion matrix data.}
Among the\CFinsert{se} classes \CFdelete{with low \(F_1\)-scores we found that there} were many \CFreplace{classes for}{representing} furry, brown\CFdelete{,} or grey animals (\emph{bear}, \emph{otter}, \emph{rabbit}, \emph{kangaroo}, and \emph{seal}).

In the ConfusionFlow matrix, we saw that these classes were frequently confused with each other \CFinsert{(see Figure~S-4 in the Supplementary Information)}.
The remaining classes with low \(F_1\)-score were related to images of young humans (i.e., \emph{boy}, \emph{girl}, and \emph{baby})\CFdelete{, which also were frequently confused with each other, but not as frequently with the remaining animal classes}.
\CFdelete{Furthermore, }ConfusionFlow's rescaling functions helped to show that the confusion values for the\CFinsert{se} three human\CFinsert{-}related classes are much higher than for the animal classes.
This means that the poor performance for animals is spread out over many more classes than in the case of humans, as the final \(F_1\)-scores were similarly bad for all classes~(\ref{task:measure-perf}).

While these findings could have been made without looking at temporal data, ConfusionFlow reveals that all these classes suffer from severe overfitting~(\ref{task:comp-perf}).
This overfitting was apparent in the precision, recall, and \(F_1\)-score plots of all classes as well as in the overall accuracy (see Figure~\CFreplace{XX}{S-2 in the Supplementary Information)}.

However, we could not find particular pairs of classes in this submatrix for which the confusion score got significantly worse over the course of training as a result of the overfitting.
Using only this class selection mechanism, we could not assess whether overfitting was a general issue or affected certain pairs of classes more severely than others.

\ifshowchanges
\begin{figure}
    \centering
    \includegraphics[width=.75\columnwidth]{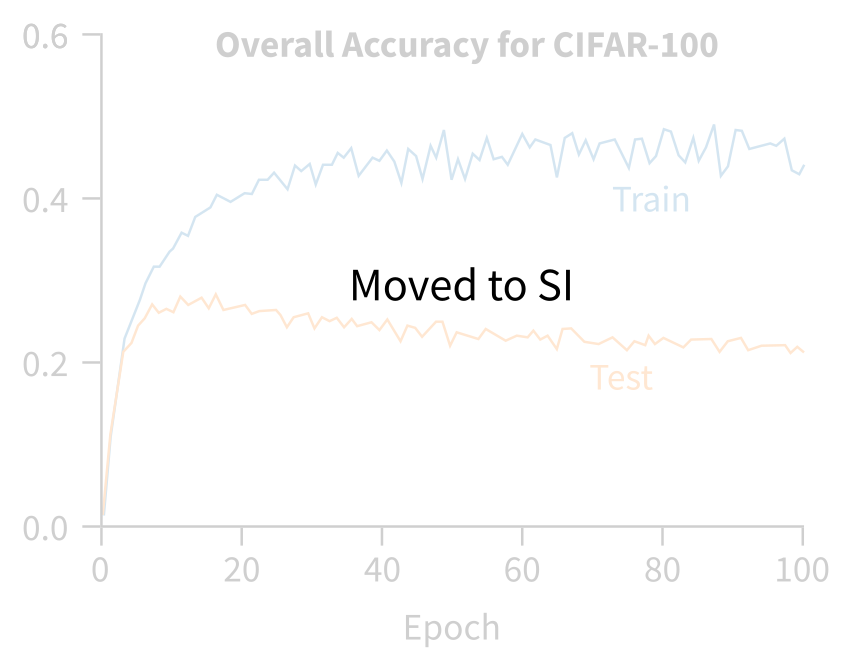}%
    \def\thefigure{XX}
    \CFdelete{\caption{Overfitting apparent in the overall accuracy for the CIFAR-100 train and test sets (plot slightly adapted from ConfusionFlow detail view).}}
    \addtocounter{figure}{-1}
    \label{fig:overfitting}
\end{figure}
\fi

We thus moved on to evaluate a second class selection mechanism.
We searched for the ten largest off-diagonal values in the confusion matrix for the test set and for the final training epoch.
In the pairs of classes describing these cells, 14~classes were represented \CFinsert{(see Figure~S-5 in the Supplementary Information)}.
Among these 14~classes are three different types of trees and four different types of flowers.
The classes \emph{boy} and \emph{girl} are part of the list again, which matched our previous findings.
\CFdelete{Interestingly, the worst classes in terms of particular confusion values also include the triple \emph{bridge}, \emph{castle}, and \emph{house}.
A~look at the ConfusionFlow matrix confirms that, while the confusion between \emph{castle} and \emph{house} is highest, the other two pairs (\emph{bridge} vs.~\emph{house} and \emph{bridge} vs.~\emph{castle}) are almost as often confused with each other.}
The temporal aspect of ConfusionFlow reveals that the performance of the tree-related classes does not suffer as badly from overfitting as most other classes~(\ref{task:comp-prog}).

Had we not chosen the CIFAR-100 dataset for its superclass structure, these results from the class selection strategy would have hinted strongly at the presence of \CFreplace{some sort of}{a} class hierarchy.
\CFdelete{For many practical problems similar class hierarchies might exist.}
While ConfusionFlow was not designed specifically for analyzing class hierarchies, it can help to build some intuition about their possible existence.

We \CFdelete{p}resumed our evaluation by looking at the ConfusionFlow matrix for all 20~superclasses \CFinsert{(see Figure~S-3 in the Supplementary Information)}.
We realized from the precision, recall, and \(F_1\)-charts that the performance on the test set for all superclasses except for \emph{trees} gets worse compared to the performance on the training set~(\ref{task:comp-prog}).
This is in line with the results for the individual tree classes.
It seems that the network has \CFinsert{such} a hard time to distinguish between different types of trees, \CFdelete{so }that it is not capable of overfitting.
From looking \CFinsert{exclusively} at \CFdelete{only }the superclass confusion matrix, it would seem like this is an advantage, but obviously this behavior is only caused by generally high confusion \emph{within} the superclass.

There are some superclass pairings that have large confusion values \emph{between} superclasses\CFreplace{, such as \emph{aquatic mammals} vs.~\emph{fish}, which is not surprising as the former includes not only images of beavers, otters, and seals but also of dolphins and whales.
For some cells of the ConfusionFlow matrix we found }{with temporal characteristics} that hint\CFdelete{s} at \CFdelete{specific }overfitting problems.
\CFreplace{C}{In particular, c}onfusion values for the two vehicle classes \emph{vehicles~1} and \emph{vehicles~2} increase over time\CFdelete{, as do the values for \emph{flowers} vs.~\emph{fruits\slash vegetables}, and \emph{large carnivores} vs.~\emph{large omnivores\slash herbivores}}~(\ref{task:comp-prog}).

\begin{figure}
    \centering
    \includegraphics[width=.95\columnwidth]{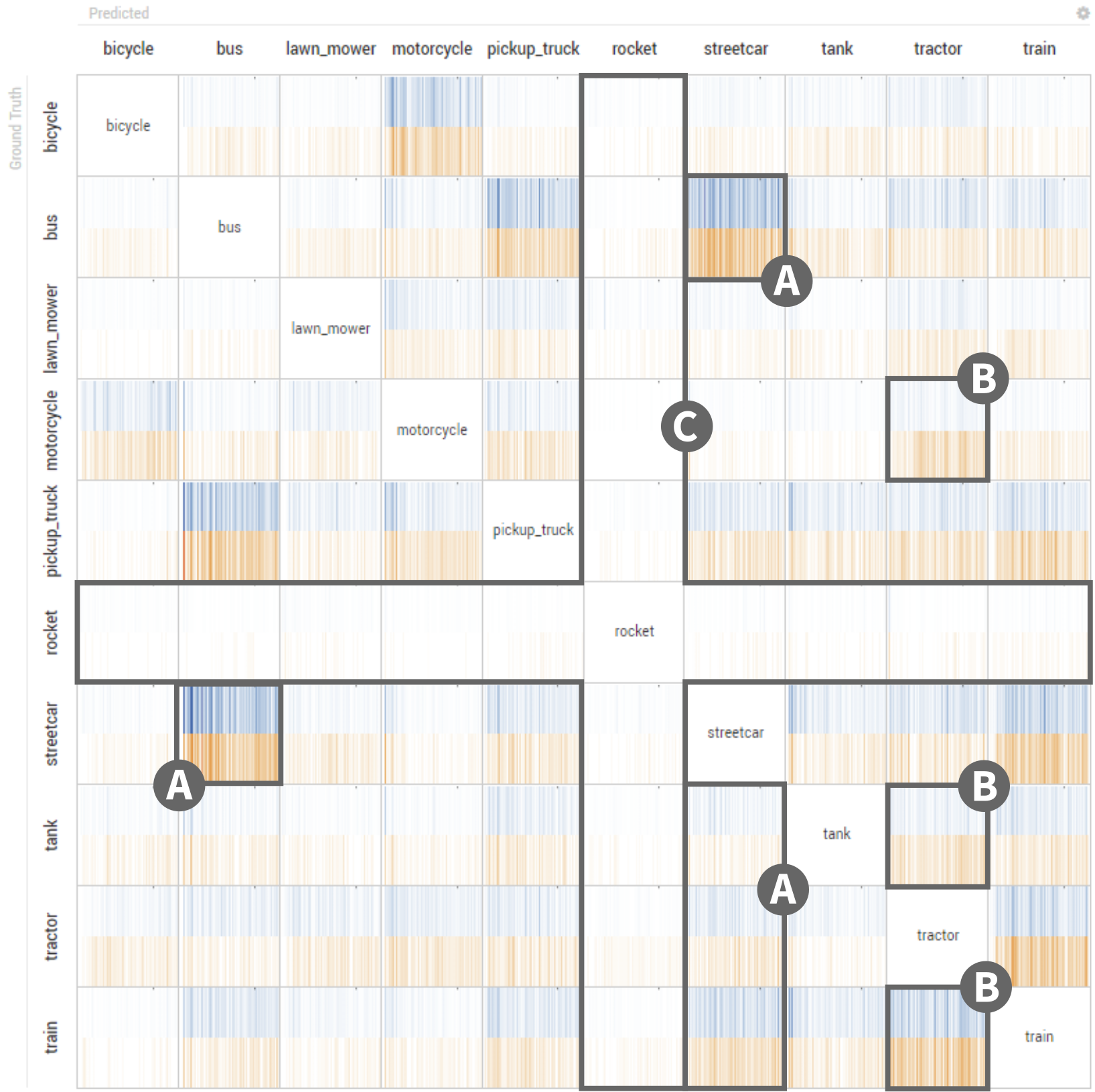}
    \caption{ConfusionFlow matrix for the ten classes making up the \emph{vehicle~1} and \emph{vehicle~2} superclasses \CFreplace{in}{for the train (\protect\colorswatch{CFblue}) and test folds (\protect\colorswatch{CForange}) of} CIFAR-100.
    \emph{Streetcar} is more often confused with \emph{bus} than with classes from its own vehicle superclass~(a). Performance for \emph{tractor} images  suffers particularly from overfitting~(b). \CFreplace{And, u}{U}nsurprisingly, \emph{rocket} seems to be hardly confused with any other vehicle~(c), even though too long training causes some tractors and buses to \CFreplace{\enquote{become}}{be misclassified as} rockets.
    }
    \label{fig:eval-within-superclass}
\end{figure}

With the insights gained from the class aggregation we could now go back and explore the fine grained classes again.
\CFreplace{In particular, w}{W}e looked at the ten classes making up the superclasses \emph{vehicles~1} (\emph{bicycle}, \emph{bus}, \emph{motorcycle}, \emph{pickup truck}, and \emph{train}) and \emph{vehicles~2} (\emph{lawn-mower}, \emph{rocket}, \emph{streetcar}, \emph{tank}, and \emph{tractor}).
The ConfusionFlow visualization for these ten classes is shown in Figure~\ref{fig:eval-within-superclass}.
\CFdelete{This analysis makes the weaknesses of only relying on the superclass approach apparent.}
The confusion values for the pairs of classes within \emph{vehicles~1} are generally much higher than those for pairs of classes within \emph{vehicles~2} and pairs of classes across the two superclasses.
A~strong exception to this rule is class pairings featuring the \emph{streetcar} class (see Figure~\ref{fig:eval-within-superclass}\annotref{a}), which hints at a flawed definition of superclasses.
It seems \CFreplace{that swapping the}{appropriate to swap} the \CFinsert{super}class\CFdelete{es} \CFinsert{memberships of} \emph{train} and \emph{streetcar}\CFdelete{ would be more appropriate, at least with regard to the presently studied classifier}.

Again, the temporal aspect of ConfusionFlow was helpful in assessing problems caused by overfitting.
In particular, the performance of \emph{tractor} gets worse over time~(\ref{task:comp-prog}), which is most likely related to confusion with \emph{train} and \emph{tank} (see Figure~\ref{fig:eval-within-superclass}\annotref{b}.
Interestingly, while the network is generally good at distinguishing \emph{rockets} from other vehicles, too long training causes some \emph{tractor} and \emph{bus} images to be classified as \emph{rockets} (Figure~\ref{fig:eval-within-superclass}\annotref{c}).
These temporal anomalies (\ref{task:detect-problem}) would have been hard to detect from a single \enquote{snapshot} of the sparse confusion matrix.

\CFdelete{%
\subsubsection*{Reflections on Scalability}

The scalability issues with ConfusionFlow are, as expected, closely related to the scalability issues of the confusion matrix.
Larger confusion matrices necessarily become more sparse or have smaller values in the off-diagonal.
The class selection strategy can reveal interesting information, but suffers from the sparseness of the matrix, no matter which selection scheme is used.
The class aggregation mechanism, on the other hand, is more suitable for giving a quick overview without completely disregarding any classes.
Instead, combinations of classes from the same superclass are disregarded, which hides many of the worst classification errors.

In our opinion, the confusion matrix for many classes, and thus also the ConfusionFlow visualization, is best explored in a three step fashion.
First, some particularly problematic classes need to be identified externally by means of some numeric evaluation.
This subset of classes can be viewed in ConfusionFlow to partially assess the possibility of a class hierarchy.
However, ConfusionFlow was not designed for this task.
To fully analyze a possible class hierarchy, more specialized tools such as Blocks by Alsallakh et al.~\cite{alsallakh_convolutional_2018} could be applied.
If a class hierarchy is established, the ConfusionFlow visualization can be applied to the superclasses, yielding high-level temporal information.
Finally, interesting findings from the superclass performance analysis have to be re-evaluated on the more detailed level of classes, in particular to better assess the within-superclass confusions.

On possible scalability issue not directly related to the confusion matrix is a general information overload caused by the high number of time series that are shown in the same view.
To make ConfusionFlow more suitable for datsets with many classes, more advanced focus~+~con\-text techniques would need to be added, similar to those provided by tools such as LiveRAC~\cite{mclachlan_liverac_2008}.

All in all, the scalability issues and mitigation mechanisms confirm our view that ConfusionFlow can only serve as one step in a more elaborate and specialized workflow for classifier analysis.
We discuss this issue in more detail in Section~XX.
However, in the context of many classes, ConfusionFlow's focus on temporal analysis (\ref{hl-task:temp}) offers one particular advantage:
users can judge whether high confusion scores are only side-effects of the noise in the confusion matrix (tasks \ref{task:detect-problem} and \ref{task:relate-problem}).
This noise is closely related to the typical sparsity of large confusion matrices.
It would be impossible to identify temporally stable problems from only a discrete, single epoch \enquote{snapshot} of the confusion matrix.}

\CFinsert{%
\subsection{Use Case: Neural Network Pruning}
\label{sec:case-study-pruning}

Neural networks are often heavily over-parameterized and contain millions of weights whose values are irrelevant for the network's performance.
One technique for removing redundant parameters from a neural network is called \emph{pruning}.
During learning, connections in the network are successively removed (pruned) according to different selection strategies.
Successful pruning results in a compressed model that retains its accuracy while requiring fewer computations and less memory.
This facilitates deployment in embedded systems ~\cite{han_learning_2015} and sometimes even leads to faster learning and better generalization than the initial dense model~\cite{frankle_lottery_2018, liu_rethinking_2018}.

We examined the performance of several fully connected networks with different architectures trained on the Fashion-MNIST dataset~\cite{xiao_fashion-mnist:_2017}.
This dataset consists of grey-scale images of fashion objects organized in 10 classes (\emph{trouser}, \emph{pullover}, \emph{sandal}, etc.).
Specifically, we investigated the effects of pruning a neural network and re-initializing the resulting sparse network with the initial weights as described by Frankle and Carbin~\cite{frankle_lottery_2018}.
Using ConfusionFlow's temporal-comparative analysis features, we tried to get a better understanding of how removing certain weights affects the model's ability to distinguish between classes.


\begin{figure*}
    \centering
    \includegraphics[width=1.0\linewidth]{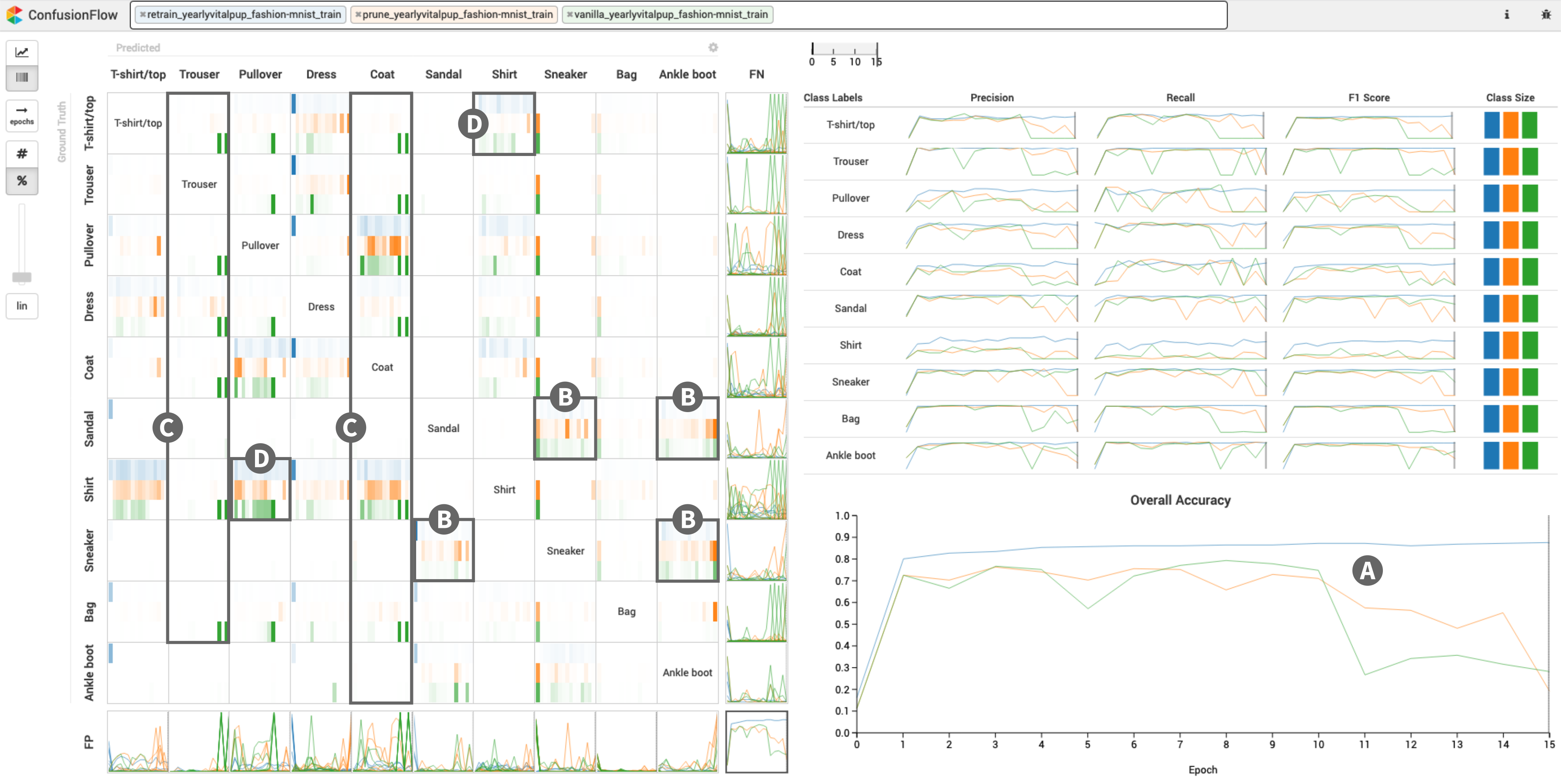}
    \caption{%
    Visual comparison of different neural network pruning strategies.
    An original network~(\protect\colorswatch{CFgreen}), a pruned network~(\protect\colorswatch{CForange}), and a re-initialized sparse network~(\protect\colorswatch{CFblue}) were trained to classify Fashion-MNIST images.
    ConfusionFlow reveals how the accuracy drop after 10 to 14 epochs~(\protect\annotref*{a}) relates to confusions for different pairs of classes~(\protect\annotref*{b}--\protect\annotref*{d}).
    The learning behavior of the re-initialized sparse network is much more stable compared to that of the other two models.
    }
\label{fig:pruningResult}
\end{figure*}

Figure~\ref{fig:pruningResult} shows the ConfusionFlow visualization for three different networks trained on the Fashion-MNIST dataset.
The original network~(\colorswatch{CFgreen}) had 6-layers, each with 200 hidden units and ReLU activation functions.
The learning rate was \(0.012\), with a batch size of \(60\).
In the second network~(\colorswatch{CForange}), 20~percent of the connections were removed at each epoch~(this approach is called \emph{online} pruning).
The sparse network resulting after 15~epochs was then re-initialized with the same weights as the original dense network, and re-trained from scratch~(\colorswatch{CFblue}).
In this network less than 4~percent of the original connections remain.

It is immediately obvious from the overall accuracy plot (Figure~\ref{fig:pruningResult}\annotref{a}) that the training of the original model~(\colorswatch{CFgreen}) fails completely after 10~epochs~(\ref{task:detect-problem}).
Training of the online-pruned network~(\colorswatch{CForange}) fails slightly later~(\ref{task:relate-problem}).
The performance of the re-initialized sparse network~(\colorswatch{CFblue}), however, remains high.
Remarkably, it even performs better than the other two networks right from the start~(\ref{task:comp-prog}).
ConfusionFlow allows relating this global~(\ref{lvl:global}) accuracy improvement to pairwise class confusions.

Inspection of the ConfusionFlow matrix shows that the confusion counts for all shoe-related pairs of classes (\emph{sneaker} vs.~\emph{sandal}, \emph{ankle boot} vs.~\emph{sneaker}, etc.) increase considerably during later epochs for the non-pruned and online-pruned networks (Figure~\ref{fig:pruningResult}\annotref{b}).
The re-initialized sparse network, on the other hand, continues to learn to better distinguish between these classes~(\ref{task:comp-prog}).
Another reason for the complete failure of the original network seems to be related to the classes \emph{trouser} and \emph{coat} (see Figure~\ref{fig:pruningResult}\annotref{c}), with extremely high FP~values for these two classes in two of the later epochs~(\ref{task:relate-problem}).

Even though the global accuracy plot showed a pronounced accuracy drop for the non-pruned and the online-pruned networks, both models retain an accuracy of about 30~percent~(\ref{task:measure-perf}).
The ConfusionFlow matrix reveals that this remaining accuracy in the later epochs is related to a better performance for the pairs \emph{shirt} vs.~\emph{T-shirt/top} and \emph{pullover} vs.~\emph{shirt} (Figure~\ref{fig:pruningResult}\annotref{d}).
The better generalization of the re-initialized sparse network across other classes comes at the cost of higher confusion values for images showing different kinds of upper body garments.

These findings demonstrate that ConfusionFlow allows a more nuanced evaluation of classifier performance, enabling the user to trace accuracy changes back to the class level.
}

\CFinsert{\subsection{Summary \& Discussion of Findings}

\subsubsection*{Visual Design}

Our collaborators found ConfusionFlow easy to use and appreciated that its visual design incorporated two visualization idioms that they were already acquainted with: the confusion matrix and temporal line charts.
The familiar layout of the confusion matrix along with the consistent color-coding of models helped our collaborators to navigate their analysis through the information-rich display.
They mentioned, however, that in the case of more classes or more models they would welcome additional focus~+~con\-text capabilities, perhaps similar to those provided by tools such as LiveRAC~\cite{mclachlan_liverac_2008}.

\subsubsection*{Comparative \& Temporal Analysis on the Class Level}

The different scenarios in our case study on active learning and our scalability evaluation revealed a general strength of the combined temporal and comparative analysis capabilities (\ref{hl-task:comp} + \ref{hl-task:temp}) in ConfusionFlow, in particular with regards to the class level (\ref{lvl:class}).
As classifiers learn, the off-diagonal elements of the confusion matrix tend to get sparse.
Temporally fine-grained learning---such as model updates after each minibatch or even after individual instance, as is the case in active learning---can result in confusion matrices of final classifiers that may not be representative of general patterns.
Our collaborators appreciated ConfusionFlow's temporal-comparative approach, as it enabled them to identify temporally stable patterns.
We found the same aspect useful in our scalability study, where the high number of classes leads to particularly sparse matrices.
Furthermore, looking at the class level reveals how likely the model is to fluctuate.
Global-level performance measures tend to hide this susceptibility of models to random or systematic changes during the training, as they average over all classes.

\subsubsection*{Scalability}

Our scalability study with the CIFAR-100 dataset showed that ConfusionFlow can be used to analyze up to about 20 classes, although for more than 15 classes, screen resolution and label placement start to become limiting factors.
It was still possible to derive insights from the 20-superclass confusion matrix, which could be used downstream in a subsequent class-level analysis.}

\subsubsection*{Bidirectional Analysis}

\CFdelete{As already stated in the introduction, developing a ML model requires a highly iterative workflow.
On its own, ConfusionFlow can only provide the tools for analyzing classification results.}
\CFreplace{As such, u}{Typically, u}sers apply ConfusionFlow after they already went through a reasoning process about a model's potential usefulness.
With enough prior knowledge about the \CFreplace{thoughts that were put into the}{motivations behind the} initial models\CFreplace{ and datasets they are applied to, the way from an analysis in ConfusionFlow back to a new model is sometimes possible without any further tools.}{, insights gained with ConfusionFlow can be directly used to \enquote{go back} and optimize these models---resulting in a \emph{bidirectional} analysis and development workflow.}

However, in most cases \CFreplace{a truly}{true} bidirectional\CFinsert{ity} \CFreplace{analysis and development workflow} requires further tools.
\CFreplace{In many cases, these tools need to be}{The three scenarios presented in this section showed that the specific requirements for a holistic analysis are} domain-specific \CFreplace{,}{and/or} model-specific (keep in mind that ConfusionFlow is fully model-agnostic)\CFreplace{ and/or offer information on an instance level~(\ref{lvl:instance})}.
\CFinsert{Additionally, in some cases instance-level information~(\ref{lvl:instance}) is required}.

\CFdelete{Systems such as LIME~\cite{ribeiro_why_2016} provide instance-based explanations of classification results specific to the ML task at hand.
In other cases, a more detailed exploration of class-relationships might be necessary, for example with regards to class hierarchies (see Section~\ref{sec:eval-scalability}, where we mention Blocks by Allsalakh et al.~\cite{alsallakh_convolutional_2018} as a possible tool for a follow-up analysis).
Again in other cases, the used data sets might have to be explored in more depth, for instance to confirm a possible distribution shift.}

\CFreplace{W}{After discussion with our collaborators, w}e see ConfusionFlow as one step in \CFreplace{this}{a} bidirectional, iterative analysis workflow\CFreplace{, and we}{. We} see its main strength in that it provides temporal and comparative insights \CFreplace{that can be used}{which can serve}  as additional mental inputs for a more \CFreplace{focused}{application-specific} analysis\CFdelete{ in specialized tools}.
\CFinsert{Examples of tools that could be used in conjunction with ConfusionFlow are LIME~\cite{ribeiro_why_2016} for instance-based explanations and Blocks by Allsalakh et al.~\cite{alsallakh_convolutional_2018} for exploring class hierarchies.}
Section~\ref{sec:instanceflow} \CFdelete{below} gives an outlook on how we plan to develop \CFreplace{a similar analysis springboard operating on the} a \CFinsert{temporal} instance\CFinsert{-}level \CFinsert{visualization that will serve as an analysis springboard similar to ConfusionFlow}.
\CFinsert{Our collaborators showed particular interest in such an instance-level tool for further continuing their analysis of selection strategies based on insights gained with ConfusionFlow.}

\CFinsert{\section{Future Work}}
\label{sec:limitations}

\subsection{Instance-Level Analysis}
\label{sec:instanceflow}

Since ConfusionFlow was specifically designed to visualize class-level information (\ref{lvl:class}), it does not \CFreplace{feature tools for analyzing the confusion on an}{enable} instance-\CFdelete{based }level \CFinsert{analysis}~(\ref{lvl:instance})\CFinsert{.} \CFdelete{over the learning process (neither regarding assignment to the wrong class, nor per-instance classification uncertainty).}
However, exploring the learning dynamics at the level of instances \CFreplace{could provide valuable additional insight into the underlying dataset and the classification behavior for individual experiments.
The detection of problematic, misclassified instances could potentially}{would} help \CFinsert{users to} discover labeling errors or outlier instances that might otherwise go unnoticed.

We are currently working on InstanceFlow, a visualization tool for addressing \CFdelete{exactly} these issues.
InstanceFlow will visualize the classification progression of individual instances throughout the training \CFinsert{in a Sankey-like fashion}.
It will allow users to filter instances by certain metrics, such as the frequency of \enquote{hopping} between classes.

Especially for neural network classifiers, linking this instance-level class-confusion to a feature-based detail view could further improve the understanding of the learning behavior\CFdelete{ of the classifier}.
Depending on the supported model architectures, this detail view could build upon previous work by Olah et al.~\cite{olah_building_2018} or work regarding activation visualization~\cite{kahng_activis:_2018, pezzotti_deepeyes:_2018}.

\subsection{Multi-Label Classification}

Multi-label classification is an ML problem, in which multiple class labels can be assigned to instances.
In principle, ConfusionFlow can be extended to visualize the confusion of multi-label classifiers, by using combinations of class labels instead of single classes along each of the matrix axes.
However, since the number of possible combinations grows quickly with the number of overall classes, aggregation methods would need to be incorporated in the workflow.
\CFreplace{Thus, in case of multi-label classification, an instance-level approach seems more appropriate.}{Alternatively, the ConfusionFlow technique could be paired with instance selection strategies to visualize instance-wise multilabel confusion matrices.}

\section{Conclusion}
\label{sec:conclusion}

In this paper we introduced ConfusionFlow, a novel tool for visualizing and exploring the temporal progression of classifier confusion.
ConfusionFlow combines a visualization of the confusion matrix over time with charts for global and per-class performance metrics.
We evaluated the usefulness of ConfusionFlow's interactive exploration capabilities by means of a case study on \CFdelete{the performance of }instance selection strategies in active learning.
\CFreplace{We f}{F}urthermore\CFinsert{, we} analyzed ConfusionFlow's scalability \CFreplace{issues and \CFinsert{discussed} possible mitigation mechanisms for classification problems with many classes}{and presented a use case in the context of neural network pruning}.

\CFdelete{We would like to point out that we designed }ConfusionFlow \CFinsert{was} not \CFinsert{designed} as a catch-all, standalone tool, but to be used in \CFreplace{combination}{conjunction} with other tools and visualization components.
In particular, we plan to complement ConfusionFlow's class-level information with a novel visualization tool focused on temporal observation of instance-level confusion.
However, offering model comparison and temporal training analysis at the class level, ConfusionFlow can fill an important gap in an ML workflow towards understanding and interpreting classification models.

\section*{Acknowledgments}

This work was supported in part by the State of Upper Austria (FFG 851460, Human-Interpretable Machine Learning) and the Austrian Science Fund (FWF P27975-NBL).




\bibliographystyle{abbrv-doi}
\bibliography{cf-bib}

\begin{thebibliography}{10}

\bibitem{abadi_tensorflow:_2016}
M.~Abadi, A.~Agarwal, P.~Barham, E.~Brevdo, Z.~Chen, C.~Citro, G.~S. Corrado,
  A.~Davis, J.~Dean, M.~Devin, S.~Ghemawat, I.~Goodfellow, A.~Harp, G.~Irving,
  M.~Isard, Y.~Jia, R.~Jozefowicz, L.~Kaiser, M.~Kudlur, J.~Levenberg, D.~Mane,
  R.~Monga, S.~Moore, D.~Murray, C.~Olah, M.~Schuster, J.~Shlens, B.~Steiner,
  I.~Sutskever, K.~Talwar, P.~Tucker, V.~Vanhoucke, V.~Vasudevan, F.~Viegas,
  O.~Vinyals, P.~Warden, M.~Wattenberg, M.~Wicke, Y.~Yu, and X.~Zheng.
\newblock {TensorFlow}: {Large}-{Scale} {Machine} {Learning} on {Heterogeneous}
  {Distributed} {Systems}.
\newblock {\em arXiv:1603.04467 [cs]}, 2016.

\bibitem{aigner2011visualization}
W.~Aigner, S.~Miksch, H.~Schumann, and C.~Tominski.
\newblock {\em Visualization of time-oriented data}.
\newblock Springer Science \& Business Media, 2011.

\bibitem{alsallakh_visual_2014}
B.~Alsallakh, A.~Hanbury, H.~Hauser, S.~Miksch, and A.~Rauber.
\newblock Visual {Methods} for {Analyzing} {Probabilistic} {Classification}
  {Data}.
\newblock {\em IEEE Transactions on Visualization and Computer Graphics},
  20(12):1703--1712, 2014. doi: {{%
10\hspace{.1pt}\discretionary{.}{%
}{.}\hspace{.4pt}1109\discretionary{/}{%
}{/}TVCG\hspace{.1pt}\discretionary{.}{%
}{.}\hspace{.4pt}2014\hspace{.1pt}\discretionary{.}{%
}{.}\hspace{.4pt}2346660}}


\bibitem{alsallakh_convolutional_2018}
B.~Alsallakh, A.~Jourabloo, M.~Ye, X.~Liu, and L.~Ren.
\newblock Do {Convolutional} {Neural} {Networks} {Learn} {Class} {Hierarchy}?
\newblock {\em IEEE Transactions on Visualization and Computer Graphics},
  24(1):152--162, 2018. doi: {{%
10\hspace{.1pt}\discretionary{.}{%
}{.}\hspace{.4pt}1109\discretionary{/}{%
}{/}TVCG\hspace{.1pt}\discretionary{.}{%
}{.}\hspace{.4pt}2017\hspace{.1pt}\discretionary{.}{%
}{.}\hspace{.4pt}2744683}}


\bibitem{andrienko2006exploratory}
N.~Andrienko and G.~Andrienko.
\newblock {\em Exploratory analysis of spatial and temporal data: a systematic
  approach}.
\newblock Springer Science \& Business Media, 2006.

\bibitem{bernard2015}
J.~Bernard.
\newblock {\em Exploratory search in time-oriented primary data}.
\newblock PhD thesis, Technische Universit{\"a}t, Darmstadt, 2015.

\bibitem{bernard_visualinteractive_2019}
J.~Bernard, M.~Hutter, H.~Reinemuth, H.~Pfeifer, C.~Bors, and J.~Kohlhammer.
\newblock Visual-{Interactive} {Preprocessing} of {Multivariate} {Time}
  {Series} {Data}.
\newblock {\em Computer Graphics Forum}, 38(3):401--412, 2019. doi: {{%
10\hspace{.1pt}\discretionary{.}{%
}{.}\hspace{.4pt}1111\discretionary{/}{%
}{/}cgf\hspace{.1pt}\discretionary{.}{%
}{.}\hspace{.4pt}13698}}


\bibitem{bernard_comparing_2018}
J.~Bernard, M.~Hutter, M.~Zeppelzauer, D.~Fellner, and M.~Sedlmair.
\newblock Comparing {Visual}-{Interactive} {Labeling} with {Active} {Learning}:
  {An} {Experimental} {Study}.
\newblock {\em IEEE Transactions on Visualization and Computer Graphics},
  24(1):298--308, 2018. doi: {{%
10\hspace{.1pt}\discretionary{.}{%
}{.}\hspace{.4pt}1109\discretionary{/}{%
}{/}TVCG\hspace{.1pt}\discretionary{.}{%
}{.}\hspace{.4pt}2017\hspace{.1pt}\discretionary{.}{%
}{.}\hspace{.4pt}2744818}}


\bibitem{bernard_towards_2018}
J.~Bernard, M.~Zeppelzauer, M.~Lehmann, M.~Müller, and M.~Sedlmair.
\newblock Towards {User}-{Centered} {Active} {Learning} {Algorithms}.
\newblock {\em Computer Graphics Forum}, 37(3):121--132, 2018. doi: {{%
10\hspace{.1pt}\discretionary{.}{%
}{.}\hspace{.4pt}1111\discretionary{/}{%
}{/}cgf\hspace{.1pt}\discretionary{.}{%
}{.}\hspace{.4pt}13406}}


\bibitem{bernard_vial:_2018}
J.~Bernard, M.~Zeppelzauer, M.~Sedlmair, and W.~Aigner.
\newblock {VIAL}: a unified process for visual interactive labeling.
\newblock {\em The Visual Computer}, 34(9):1189--1207, 2018. doi: {{%
10\hspace{.1pt}\discretionary{.}{%
}{.}\hspace{.4pt}1007\discretionary{/}{%
}{/}s00371\discretionary{%
}{-}{-}018\discretionary{%
}{-}{-}1500\discretionary{%
}{-}{-}3}}


\bibitem{bostock_d3:_2011}
M.~Bostock, V.~Ogievetsky, and J.~Heer.
\newblock D3: {Data}-{Driven} {Documents}.
\newblock {\em IEEE Transactions on Visualization and Computer Graphics},
  17(12):2301--2309, 2011. doi: {{%
10\hspace{.1pt}\discretionary{.}{%
}{.}\hspace{.4pt}1109\discretionary{/}{%
}{/}TVCG\hspace{.1pt}\discretionary{.}{%
}{.}\hspace{.4pt}2011\hspace{.1pt}\discretionary{.}{%
}{.}\hspace{.4pt}185}}


\bibitem{bruckner_ml-o-scope:_2014}
D.~Bruckner.
\newblock {ML}-o-{Scope}: {A} {Diagnostic} {Visualization} {System} for {Deep}
  {Machine} {Learning} {Pipelines}.
\newblock Technical report, Defense Technical Information Center, Fort Belvoir,
  VA, 2014. doi: {{%
10\hspace{.1pt}\discretionary{.}{%
}{.}\hspace{.4pt}21236\discretionary{/}{%
}{/}ADA605112}}


\bibitem{chae_visualization_2017}
J.~Chae, S.~Gao, A.~Ramanthan, C.~Steed, and G.~D. Tourassi.
\newblock Visualization for {Classification} in {Deep} {Neural} {Networks}.
\newblock In {\em Workshop on {Visual} {Analytics} for {Deep} {Learning} at
  {IEEE} {VIS}}, p.~6, 2017.

\bibitem{chung_revacnn:_2016}
S.~Chung, S.~Suh, and C.~Park.
\newblock {ReVACNN}: {Real}-{Time} {Visual} {Analytics} for {Convolutional}
  {Neural} {Network}.
\newblock In {\em {ACM} {SIGKDD} {Workshop} on {Interactive} {Data}
  {Exploration} and {Analytics} ({IDEA})}, p.~7, 2016.

\bibitem{deng-imagenet-2009}
J.~Deng, W.~Dong, R.~Socher, L.-J. Li, K.~Li, and L.~Fei-Fei.
\newblock Image{N}et: A large-scale hierarchical image database.
\newblock In {\em 2009 IEEE conference on computer vision and pattern
  recognition}, pp. 248--255, 2009.

\bibitem{fayyad1996data}
U.~Fayyad, G.~Piatetsky-Shapiro, and P.~Smyth.
\newblock From data mining to knowledge discovery in databases.
\newblock {\em AI magazine}, 17(3):37--37, 1996.

\bibitem{feifei-learning-2004}
L.~Fei-Fei, R.~Fergus, and P.~Perona.
\newblock Learning generative visual models from few training examples: an
  incremental bayesian approach tested on 101 object categories.
\newblock In {\em 2004 Conference on Computer Vision and Pattern Recognition
  (CVPR) Workshop}, 2004.

\bibitem{frankle_lottery_2018}
J.~Frankle and M.~Carbin.
\newblock The {Lottery} {Ticket} {Hypothesis}: {Finding} {Sparse}, {Trainable}
  {Neural} {Networks}.
\newblock {\em arXiv:1803.03635 [cs]}, 2018.

\bibitem{FU2011164}
T.-c. Fu.
\newblock A review on time series data mining.
\newblock {\em Engineering Applications of Artificial Intelligence}, 24(1):164
  -- 181, 2011. doi: {{%
10\hspace{.1pt}\discretionary{.}{%
}{.}\hspace{.4pt}1016\discretionary{/}{%
}{/}j\hspace{.1pt}\discretionary{.}{%
}{.}\hspace{.4pt}engappai\hspace{.1pt}\discretionary{.}{%
}{.}\hspace{.4pt}2010\hspace{.1pt}\discretionary{.}{%
}{.}\hspace{.4pt}09\hspace{.1pt}\discretionary{.}{%
}{.}\hspace{.4pt}007}}


\bibitem{gleicher_visual_2011}
M.~Gleicher, D.~Albers, R.~Walker, I.~Jusufi, C.~D. Hansen, and J.~C. Roberts.
\newblock Visual comparison for information visualization.
\newblock {\em Information Visualization}, 10(4):289 --309, 2011. doi: {{%
10\hspace{.1pt}\discretionary{.}{%
}{.}\hspace{.4pt}1177\discretionary{/}{%
}{/}1473871611416549}}


\bibitem{glorot_domain_2011}
X.~Glorot, A.~Bordes, and Y.~Bengio.
\newblock Domain {Adaptation} for {Large}-{Scale} {Sentiment} {Classification}:
  {A} {Deep} {Learning} {Approach}.
\newblock In {\em Proceedings of the 28th {International} {Conference} on
  {Machine} {Learning}}, p.~8, 2011.

\bibitem{gogolou_comparing_2019}
A.~Gogolou, T.~Tsandilas, T.~Palpanas, and A.~Bezerianos.
\newblock Comparing similarity perception in time series visualizations.
\newblock {\em {IEEE} Transactions on Visualization and Computer Graphics},
  25:523--533, 2019.

\bibitem{griffin-caltech-2007}
G.~Griffin, A.~Holub, and P.~Perona.
\newblock The caltech 256.
\newblock Technical report, Caltech, 2007.

\bibitem{han_learning_2015}
S.~Han, J.~Pool, J.~Tran, and W.~Dally.
\newblock Learning both weights and connections for efficient neural networks.
\newblock In {\em Advances in {N}eural {I}nformation {P}rocessing {S}ystems 28
  ({NIPS '15})}, pp. 1135--1143, 2015.

\bibitem{he_deep_2016}
K.~He, X.~Zhang, S.~Ren, and J.~Sun.
\newblock Deep {Residual} {Learning} for {Image} {Recognition}.
\newblock In {\em 2016 {IEEE} {Conference} on {Computer} {Vision} and {Pattern}
  {Recognition} ({CVPR})}, pp. 770--778. IEEE, Las Vegas, NV, USA, 2016. doi:
  {{%
10\hspace{.1pt}\discretionary{.}{%
}{.}\hspace{.4pt}1109\discretionary{/}{%
}{/}CVPR\hspace{.1pt}\discretionary{.}{%
}{.}\hspace{.4pt}2016\hspace{.1pt}\discretionary{.}{%
}{.}\hspace{.4pt}90}}


\bibitem{hohman_shapeshop:_2017}
F.~Hohman, N.~Hodas, and D.~H. Chau.
\newblock {ShapeShop}: {Towards} {Understanding} {Deep} {Learning}
  {Representations} via {Interactive} {Experimentation}.
\newblock In {\em Proceedings of the 2017 {CHI} {Conference} {Extended}
  {Abstracts} on {Human} {Factors} in {Computing} {Systems} ({CHI} {EA} '17)},
  pp. 1694--1699, 2017. doi: {{%
10\hspace{.1pt}\discretionary{.}{%
}{.}\hspace{.4pt}1145\discretionary{/}{%
}{/}3027063\hspace{.1pt}\discretionary{.}{%
}{.}\hspace{.4pt}3053103}}


\bibitem{hohman_visual_2018}
F.~Hohman, M.~Kahng, R.~Pienta, and D.~H. Chau.
\newblock Visual {Analytics} in {Deep} {Learning}: {An} {Interrogative}
  {Survey} for the {Next} {Frontiers}.
\newblock {\em IEEE Transactions on Visualization and Computer Graphics},
  25(8):2674--2693, 2018. doi: {{%
10\hspace{.1pt}\discretionary{.}{%
}{.}\hspace{.4pt}1109\discretionary{/}{%
}{/}TVCG\hspace{.1pt}\discretionary{.}{%
}{.}\hspace{.4pt}2018\hspace{.1pt}\discretionary{.}{%
}{.}\hspace{.4pt}2843369}}


\bibitem{kahng_activis:_2018}
M.~Kahng, P.~Y. Andrews, A.~Kalro, and D.~H.~. Chau.
\newblock {ActiVis}: {Visual} {Exploration} of {Industry}-{Scale} {Deep}
  {Neural} {Network} {Models}.
\newblock {\em IEEE Transactions on Visualization and Computer Graphics},
  24(1):88--97, 2018. doi: {{%
10\hspace{.1pt}\discretionary{.}{%
}{.}\hspace{.4pt}1109\discretionary{/}{%
}{/}TVCG\hspace{.1pt}\discretionary{.}{%
}{.}\hspace{.4pt}2017\hspace{.1pt}\discretionary{.}{%
}{.}\hspace{.4pt}2744718}}


\bibitem{kahng_gan_2019}
M.~Kahng, N.~Thorat, D.~H.~P. Chau, F.~B. Viegas, and M.~Wattenberg.
\newblock {GAN} {Lab}: {Understanding} {Complex} {Deep} {Generative} {Models}
  using {Interactive} {Visual} {Experimentation}.
\newblock {\em IEEE Transactions on Visualization and Computer Graphics},
  25(1):1--11, 2019. doi: {{%
10\hspace{.1pt}\discretionary{.}{%
}{.}\hspace{.4pt}1109\discretionary{/}{%
}{/}TVCG\hspace{.1pt}\discretionary{.}{%
}{.}\hspace{.4pt}2018\hspace{.1pt}\discretionary{.}{%
}{.}\hspace{.4pt}2864500}}


\bibitem{kapoor_interactive_2010}
A.~Kapoor, B.~Lee, D.~Tan, and E.~Horvitz.
\newblock Interactive optimization for steering machine classification.
\newblock In {\em Proceedings of the 28th international conference on {Human}
  factors in computing systems ({CHI} '10)}, p. 1343. ACM Press, 2010. doi: {{%
10\hspace{.1pt}\discretionary{.}{%
}{.}\hspace{.4pt}1145\discretionary{/}{%
}{/}1753326\hspace{.1pt}\discretionary{.}{%
}{.}\hspace{.4pt}1753529}}


\bibitem{krizhevsky_learning_2009}
A.~Krizhevsky.
\newblock Learning {Multiple} {Layers} of {Features} from {Tiny} {Images}.
\newblock Technical Report Vol. 1, No. 4, University of Toronto, 2009.

\bibitem{krizhevsky_imagenet_2017}
A.~Krizhevsky, I.~Sutskever, and G.~E. Hinton.
\newblock {ImageNet} classification with deep convolutional neural networks.
\newblock {\em Communications of the ACM}, 60(6):84--90, 2017. doi: {{%
10\hspace{.1pt}\discretionary{.}{%
}{.}\hspace{.4pt}1145\discretionary{/}{%
}{/}3065386}}


\bibitem{lecun_gradient-based_1998}
Y.~LeCun, L.~Bottou, Y.~Bengio, and P.~Haffner.
\newblock Gradient-based learning applied to document recognition.
\newblock {\em Proceedings of the IEEE}, 86(11):2278--2324, 1998. doi: {{%
10\hspace{.1pt}\discretionary{.}{%
}{.}\hspace{.4pt}1109\discretionary{/}{%
}{/}5\hspace{.1pt}\discretionary{.}{%
}{.}\hspace{.4pt}726791}}


\bibitem{liu_deeptracker:_2018}
D.~Liu, W.~Cui, K.~Jin, Y.~Guo, and H.~Qu.
\newblock {DeepTracker}: Visualizing the training process of convolutional
  neural networks.
\newblock {\em {ACM} {T}ransactions on {I}ntelligent {S}ystems and {T}echnology
  ({TIST})}, 10(1):1--25, 2018.

\bibitem{liu_rethinking_2018}
Z.~Liu, M.~Sun, T.~Zhou, G.~Huang, and T.~Darrell.
\newblock Rethinking the {Value} of {Network} {Pruning}.
\newblock {\em arXiv:1810.05270 [cs, stat]}, 2018.

\bibitem{mayr_large-scale_2018}
A.~Mayr, G.~Klambauer, T.~Unterthiner, M.~Steijaert, J.~K. Wegner,
  H.~Ceulemans, D.-A. Clevert, and S.~Hochreiter.
\newblock Large-scale comparison of machine learning methods for drug target
  prediction on {ChEMBL}.
\newblock {\em Chemical Science}, 9(24):5441--5451, 2018. doi: {{%
10\hspace{.1pt}\discretionary{.}{%
}{.}\hspace{.4pt}1039\discretionary{/}{%
}{/}C8SC00148K}}


\bibitem{mclachlan_liverac_2008}
P.~McLachlan, T.~Munzner, E.~Koutsofios, and S.~North.
\newblock Live{RAC}: interactive visual exploration of system management
  time-series data.
\newblock In {\em Proceedings of the {SIGCHI} Conference on Human Factors in
  Computing Systems ({CHI '08})}, pp. 1483--1492, 2008.

\bibitem{MIKSCH2014286}
S.~Miksch and W.~Aigner.
\newblock A matter of time: Applying a data–users–tasks design triangle to
  visual analytics of time-oriented data.
\newblock {\em Computers \& Graphics}, 38:286 -- 290, 2014. doi: {{%
10\hspace{.1pt}\discretionary{.}{%
}{.}\hspace{.4pt}1016\discretionary{/}{%
}{/}j\hspace{.1pt}\discretionary{.}{%
}{.}\hspace{.4pt}cag\hspace{.1pt}\discretionary{.}{%
}{.}\hspace{.4pt}2013\hspace{.1pt}\discretionary{.}{%
}{.}\hspace{.4pt}11\hspace{.1pt}\discretionary{.}{%
}{.}\hspace{.4pt}002}}


\bibitem{ming_understanding_2017}
Y.~Ming, S.~Cao, R.~Zhang, Z.~Li, Y.~Chen, Y.~Song, and H.~Qu.
\newblock Understanding {Hidden} {Memories} of {Recurrent} {Neural} {Networks}.
\newblock In {\em 2017 {IEEE} {Conference} on {Visual} {Analytics} {Science}
  and {Technology} ({VAST})}, pp. 13--24, 2017. doi: {{%
10\hspace{.1pt}\discretionary{.}{%
}{.}\hspace{.4pt}1109\discretionary{/}{%
}{/}VAST\hspace{.1pt}\discretionary{.}{%
}{.}\hspace{.4pt}2017\hspace{.1pt}\discretionary{.}{%
}{.}\hspace{.4pt}8585721}}


\bibitem{nogueira_dos_santos_deep_2014}
C.~Nogueira~dos Santos and M.~Gatti.
\newblock Deep {Convolutional} {Neural} {Networks} for {Sentiment} {Analysis}
  of {Short} {Texts}.
\newblock In {\em Proceedings of the 25th {International} {Conference} on
  {Computational} {Linguistics} ({COLING} 2014)}, pp. 69--78. Dublin, Ireland,
  2014.

\bibitem{olah_building_2018}
C.~Olah, A.~Satyanarayan, I.~Johnson, S.~Carter, L.~Schubert, K.~Ye, and
  A.~Mordvintsev.
\newblock The {Building} {Blocks} of {Interpretability}.
\newblock {\em Distill}, 3(3):e10, 2018. doi: {{%
10\hspace{.1pt}\discretionary{.}{%
}{.}\hspace{.4pt}23915\discretionary{/}{%
}{/}distill\hspace{.1pt}\discretionary{.}{%
}{.}\hspace{.4pt}00010}}


\bibitem{pezzotti_deepeyes:_2018}
N.~Pezzotti, T.~Hollt, J.~Van~Gemert, B.~P. Lelieveldt, E.~Eisemann, and
  A.~Vilanova.
\newblock {DeepEyes}: {Progressive} {Visual} {Analytics} for {Designing} {Deep}
  {Neural} {Networks}.
\newblock {\em IEEE Transactions on Visualization and Computer Graphics},
  24(1):98--108, 2018. doi: {{%
10\hspace{.1pt}\discretionary{.}{%
}{.}\hspace{.4pt}1109\discretionary{/}{%
}{/}TVCG\hspace{.1pt}\discretionary{.}{%
}{.}\hspace{.4pt}2017\hspace{.1pt}\discretionary{.}{%
}{.}\hspace{.4pt}2744358}}


\bibitem{recht_cifar-10_2018}
B.~Recht, R.~Roelofs, L.~Schmidt, and V.~Shankar.
\newblock Do {CIFAR}-10 {Classifiers} {Generalize} to {CIFAR}-10?
\newblock {\em arXiv:1806.00451 [cs, stat]}, 2018.

\bibitem{ren_squares:_2017}
D.~Ren, S.~Amershi, B.~Lee, J.~Suh, and J.~D. Williams.
\newblock Squares: {Supporting} {Interactive} {Performance} {Analysis} for
  {Multiclass} {Classifiers}.
\newblock {\em IEEE Transactions on Visualization and Computer Graphics},
  23(1):61--70, 2017. doi: {{%
10\hspace{.1pt}\discretionary{.}{%
}{.}\hspace{.4pt}1109\discretionary{/}{%
}{/}TVCG\hspace{.1pt}\discretionary{.}{%
}{.}\hspace{.4pt}2016\hspace{.1pt}\discretionary{.}{%
}{.}\hspace{.4pt}2598828}}


\bibitem{ribeiro_why_2016}
M.~T. Ribeiro, S.~Singh, and C.~Guestrin.
\newblock "{Why} {Should} {I} {Trust} {You}?": {Explaining} the {Predictions}
  of {Any} {Classifier}.
\newblock In {\em Proceedings of the 22nd {ACM SIGKDD} {I}international
  {C}onference on {K}nowledge {D}iscovery and {D}ata {M}ining (KDD '16)}, pp.
  1135--1144, 2016.

\bibitem{settles_active_2012}
B.~Settles.
\newblock {\em Active {Learning}}.
\newblock Synthesis {Lectures} on {Artificial} {Intelligence} and {Machine}
  {Learning}. Morgan \& Claypool Publishers, 2012.

\bibitem{simonyan_deep_2013}
K.~Simonyan, A.~Vedaldi, and A.~Zisserman.
\newblock Deep {Inside} {Convolutional} {Networks}: {Visualising} {Image}
  {Classification} {Models} and {Saliency} {Maps}.
\newblock {\em arXiv:1312.6034 [cs]}, 2013.

\bibitem{smilkov_direct-manipulation_2017}
D.~Smilkov, S.~Carter, D.~Sculley, F.~B. Viégas, and M.~Wattenberg.
\newblock Direct-{Manipulation} {Visualization} of {Deep} {Networks}.
\newblock {\em arXiv:1708.03788 [cs, stat]}, 2017.

\bibitem{socher_recursive_2013}
R.~Socher, A.~Perelygin, J.~Wu, J.~Chuang, C.~D. Manning, A.~Ng, and C.~Potts.
\newblock Recursive {Deep} {Models} for {Semantic} {Compositionality} {Over} a
  {Sentiment} {Treebank}.
\newblock In {\em Conference on {Empirical} {Methods} in {Natural} {Language}
  {Processing}}, pp. 1631--1642, 2013.

\bibitem{sokolova_systematic_2009}
M.~Sokolova and G.~Lapalme.
\newblock A systematic analysis of performance measures for classification
  tasks.
\newblock {\em Information Processing \& Management}, 45(4):427--437, 2009.
  doi: {{%
10\hspace{.1pt}\discretionary{.}{%
}{.}\hspace{.4pt}1016\discretionary{/}{%
}{/}j\hspace{.1pt}\discretionary{.}{%
}{.}\hspace{.4pt}ipm\hspace{.1pt}\discretionary{.}{%
}{.}\hspace{.4pt}2009\hspace{.1pt}\discretionary{.}{%
}{.}\hspace{.4pt}03\hspace{.1pt}\discretionary{.}{%
}{.}\hspace{.4pt}002}}


\bibitem{swihart_lasagna_2010}
B.~J. Swihart, B.~Caffo, B.~D. James, M.~Strand, B.~S. Schwartz, and N.~M.
  Punjabi.
\newblock Lasagna {Plots}: {A} {Saucy} {Alternative} to {Spaghetti} {Plots}.
\newblock {\em Epidemiology}, 21(5):621--625, 2010. doi: {{%
10\hspace{.1pt}\discretionary{.}{%
}{.}\hspace{.4pt}1097\discretionary{/}{%
}{/}EDE\hspace{.1pt}\discretionary{.}{%
}{.}\hspace{.4pt}0b013e3181e5b06a}}


\bibitem{talbot_ensemblematrix:_2009}
J.~Talbot, B.~Lee, A.~Kapoor, and D.~S. Tan.
\newblock {EnsembleMatrix}: {Interactive} {Visualization} to {Support}
  {Machine} {Learning} with {Multiple} {Classifiers}.
\newblock In {\em Proceedings of the {SIGCHI} {Conference} on {Human} {Factors}
  in {Computing} {Systems} ({CHI} '09)}, pp. 1283--1292, 2009. doi: {{%
10\hspace{.1pt}\discretionary{.}{%
}{.}\hspace{.4pt}1145\discretionary{/}{%
}{/}1518701\hspace{.1pt}\discretionary{.}{%
}{.}\hspace{.4pt}1518895}}


\bibitem{van_den_elzen_baobabview:_2011}
S.~van~den Elzen and J.~J. van Wijk.
\newblock {BaobabView}: {Interactive} construction and analysis of decision
  trees.
\newblock In {\em 2011 {IEEE} {Conference} on {Visual} {Analytics} {Science}
  and {Technology} ({VAST})}, pp. 151--160. IEEE, Providence, RI, USA, 2011.
  doi: {{%
10\hspace{.1pt}\discretionary{.}{%
}{.}\hspace{.4pt}1109\discretionary{/}{%
}{/}VAST\hspace{.1pt}\discretionary{.}{%
}{.}\hspace{.4pt}2011\hspace{.1pt}\discretionary{.}{%
}{.}\hspace{.4pt}6102453}}


\bibitem{vanschoren_openml_2014}
J.~Vanschoren, J.~N. van Rijn, B.~Bischl, and L.~Torgo.
\newblock {OpenML}: networked science in machine learning.
\newblock {\em ACM SIGKDD Explorations Newsletter}, 15(2):49--60, 2014. doi:
  {{%
10\hspace{.1pt}\discretionary{.}{%
}{.}\hspace{.4pt}1145\discretionary{/}{%
}{/}2641190\hspace{.1pt}\discretionary{.}{%
}{.}\hspace{.4pt}2641198}}


\bibitem{wang_dqnviz:_2019}
J.~Wang, L.~Gou, H.-W. Shen, and H.~Yang.
\newblock {DQNViz}: {A} {Visual} {Analytics} {Approach} to {Understand} {Deep}
  {Q}-{Networks}.
\newblock {\em IEEE Transactions on Visualization and Computer Graphics},
  25(1):288--298, 2019. doi: {{%
10\hspace{.1pt}\discretionary{.}{%
}{.}\hspace{.4pt}1109\discretionary{/}{%
}{/}TVCG\hspace{.1pt}\discretionary{.}{%
}{.}\hspace{.4pt}2018\hspace{.1pt}\discretionary{.}{%
}{.}\hspace{.4pt}2864504}}


\bibitem{wang_ganviz:_2018}
J.~Wang, L.~Gou, H.~Yang, and H.-W. Shen.
\newblock {GANViz}: {A} {Visual} {Analytics} {Approach} to {Understand} the
  {Adversarial} {Game}.
\newblock {\em IEEE Transactions on Visualization and Computer Graphics},
  24(6):1905--1917, 2018. doi: {{%
10\hspace{.1pt}\discretionary{.}{%
}{.}\hspace{.4pt}1109\discretionary{/}{%
}{/}TVCG\hspace{.1pt}\discretionary{.}{%
}{.}\hspace{.4pt}2018\hspace{.1pt}\discretionary{.}{%
}{.}\hspace{.4pt}2816223}}


\bibitem{google_whatif}
J.~Wexler.
\newblock The what-if tool: Code-free probing of machine learning models, 2018.
\newblock Accessed: 2020-02-14.

\bibitem{wongsuphasawat_visualizing_2018}
K.~Wongsuphasawat, D.~Smilkov, J.~Wexler, J.~Wilson, D.~Mane, D.~Fritz,
  D.~Krishnan, F.~B. Viegas, and M.~Wattenberg.
\newblock Visualizing {Dataflow} {Graphs} of {Deep} {Learning} {Models} in
  {TensorFlow}.
\newblock {\em IEEE Transactions on Visualization and Computer Graphics},
  24(1):1--12, 2018. doi: {{%
10\hspace{.1pt}\discretionary{.}{%
}{.}\hspace{.4pt}1109\discretionary{/}{%
}{/}TVCG\hspace{.1pt}\discretionary{.}{%
}{.}\hspace{.4pt}2017\hspace{.1pt}\discretionary{.}{%
}{.}\hspace{.4pt}2744878}}


\bibitem{wu_sampling_2006}
Y.~Wu, I.~Kozintsev, J.~Bouguet, and C.~Dulong.
\newblock Sampling {Strategies} for {Active} {Learning} in {Personal} {Photo}
  {Retrieval}.
\newblock In {\em 2006 {IEEE} {International} {Conference} on {Multimedia} and
  {Expo}}, pp. 529--532, 2006. doi: {{%
10\hspace{.1pt}\discretionary{.}{%
}{.}\hspace{.4pt}1109\discretionary{/}{%
}{/}ICME\hspace{.1pt}\discretionary{.}{%
}{.}\hspace{.4pt}2006\hspace{.1pt}\discretionary{.}{%
}{.}\hspace{.4pt}262442}}


\bibitem{xiao_fashion-mnist:_2017}
H.~Xiao, K.~Rasul, and R.~Vollgraf.
\newblock Fashion-{MNIST}: a {Novel} {Image} {Dataset} for {Benchmarking}
  {Machine} {Learning} {Algorithms}.
\newblock {\em arXiv:1708.07747 [cs, stat]}, 2017.

\bibitem{zeng_cnncomparator:_2017}
H.~Zeng, H.~Haleem, X.~Plantaz, N.~Cao, and H.~Qu.
\newblock {CNNComparator}: {Comparative} {Analytics} of {Convolutional}
  {Neural} {Networks}.
\newblock {\em arXiv:1710.05285 [cs]}, 2017.

\bibitem{zhang_manifold:_2019}
J.~Zhang, Y.~Wang, P.~Molino, L.~Li, and D.~S. Ebert.
\newblock Manifold: {A} {Model}-{Agnostic} {Framework} for {Interpretation} and
  {Diagnosis} of {Machine} {Learning} {Models}.
\newblock {\em IEEE Transactions on Visualization and Computer Graphics},
  25(1):364--373, 2019. doi: {{%
10\hspace{.1pt}\discretionary{.}{%
}{.}\hspace{.4pt}1109\discretionary{/}{%
}{/}TVCG\hspace{.1pt}\discretionary{.}{%
}{.}\hspace{.4pt}2018\hspace{.1pt}\discretionary{.}{%
}{.}\hspace{.4pt}2864499}}


\end{thebibliography}

%




\vspace*{-3\baselineskip}
\begin{IEEEbiography}[\biographyphoto{andreas}]{Andreas Hinterreiter}
    is a PhD student at the Institute of Computer Graphics, Johannes Kepler University (JKU) Linz.
    His research interests include dimensionality reduction and explainable AI.
    He is currently on secondment at the Biomedical Image Analysis Group at Imperial College London.
    He received his Diplomingenieur (MSc) in Technical Physics from JKU.
\end{IEEEbiography}
\vspace*{-3\baselineskip}
\begin{IEEEbiography}[\biographyphoto{peter}]{Peter Ruch}
    is a PhD student at the Institute for Machine Learning, Johannes Kepler University (JKU) Linz.
    He is interested in the application of neural networks for drug discovery, especially large-scale multi-task learning for bioactivity prediction of small molecules. He received his Master in Bioinformatics from JKU.
\end{IEEEbiography}
\vspace*{-3\baselineskip}
\begin{IEEEbiography}[\biographyphoto{holger}]{Holger Stitz}
    is a postdoctoral researcher at the  at the Institute of Computer Graphics, Johannes Kepler University (JKU) Linz.
    His main research interest is multi-attribute time-series data in the context of large networks.
    For more information see \url{http://holgerstitz.de}.
\end{IEEEbiography}
\vspace*{-3\baselineskip}
\begin{IEEEbiography}[\biographyphoto{martin}]{Martin Ennemoser}
    is a researcher and developer at Salesbeat GmbH.
    His research interests include deep learning, natural language processing and reinforcement learning.
    He received a Diplomingenieur (MSc) in Computer Science from JKU.
\end{IEEEbiography}
\vspace*{-3\baselineskip}
\begin{IEEEbiography}[\biographyphoto{juergen}]{J\"urgen Bernard}
    is a postdoctoral research fellow at the University of British Columbia, with the InfoVis Group in the computer science department.
    His research interests include visual analytics, information visualization, interactive machine learning, and visual-interactive labeling in particular.
    Bernard received a PhD degree in computer science from TU Darmstadt.
    He has received the Dirk Bartz Price for Visual Computing in Medicine
    and the Hugo-Geiger Preis for excellent Dissertations.
    He has been organizing the Workshop on Visual Analytics in Health Care (VAHC) at IEEE VIS since 2015.
\end{IEEEbiography}
\vspace*{-3\baselineskip}
\begin{IEEEbiography}[\biographyphoto{hendrik}]{Hendrik Strobelt}
    is a researcher in information visualisation, visual analytics, and machine learning at IBM Research AI, Cambridge.
    His research interests include visualization of large data sets of unstructured\slash semi-structured data, biological data, and neural network models.
    He received his PhD in computer science from Uni Konstanz.
    For more information see \url{http://hendrik.strobelt.com}.
\end{IEEEbiography}
\vspace*{-3\baselineskip}
\begin{IEEEbiography}[\biographyphoto{marc}]{Marc Streit}
    is a Full Professor of Visual Data Science at the Johannes Kepler University Linz.
    He finished his PhD at the Graz University of Technology in 2011.
    His scientific areas of interest include visualization, visual analytics, and biological data visualization.
    He is also co-founder and CEO of datavisyn, a spin-off company developing data visualization solutions for applications in pharmaceutical and biomedical R\&D.
    For more information see \url{http://marc-streit.com}.
\end{IEEEbiography}


\end{document}